\documentclass[letterpaper]{article}
\usepackage[preprint]{aaai2027}
\usepackage[hyphens]{url}
\usepackage{graphicx}
\usepackage{natbib}
\usepackage{caption}
\usepackage{amsmath}
\usepackage{amssymb}
\usepackage{float}
\usepackage{booktabs}
\usepackage{multirow}
\usepackage[table]{xcolor}

\graphicspath{{./}{Figures/}{papers/Figures/}{papers/}}

\definecolor{h2rfamilyone}{RGB}{245,250,238}
\definecolor{h2rfamilytwo}{RGB}{255,247,234}
\definecolor{h2rfamilythree}{RGB}{246,242,255}
\definecolor{h2rfamilyfour}{RGB}{255,241,244}
\definecolor{h2rfamilyfive}{RGB}{238,248,255}
\definecolor{h2rfamilysix}{RGB}{241,248,241}
\definecolor{h2rvideocond}{RGB}{229,240,249}
\definecolor{h2rframecond}{RGB}{242,235,248}
\definecolor{h2rpromptbg}{RGB}{244,248,252}
\definecolor{h2rpromptborder}{RGB}{180,198,216}
\definecolor{h2rpromptouterbg}{RGB}{252,251,247}
\definecolor{h2rpromptouterborder}{RGB}{194,198,193}
\definecolor{h2rpromptvideobg}{RGB}{246,249,252}
\definecolor{h2rpromptimagebg}{RGB}{253,249,244}
\definecolor{h2rpromptembbg}{RGB}{246,251,247}
\definecolor{h2rpromptannbg}{RGB}{249,247,252}
\definecolor{h2rpromptjudgebg}{RGB}{245,248,251}
\definecolor{h2rpromptoutputbg}{RGB}{247,251,248}
\definecolor{h2rpromptvideoaccent}{RGB}{111,151,185}
\definecolor{h2rpromptimageaccent}{RGB}{196,145,94}
\definecolor{h2rpromptembaccent}{RGB}{105,157,122}
\definecolor{h2rpromptannaccent}{RGB}{143,120,176}
\definecolor{h2rpromptjudgeaccent}{RGB}{91,126,158}
\definecolor{h2rpromptoutputaccent}{RGB}{91,143,112}

\newcommand{\promptcard}[4]{%
  \begingroup
  \setlength{\fboxsep}{6pt}%
  \noindent\fcolorbox{#1}{#2}{%
    \begin{minipage}[t]{0.92\linewidth}
      {\bfseries #3}\par
      {\color{#1}\rule{0.38\linewidth}{0.7pt}}\par\smallskip
      {\scriptsize\ttfamily\raggedright #4\par}
    \end{minipage}%
  }%
  \endgroup
}

\title{H2R-Bench: Benchmarking Human-to-Robot Manipulation Video Generation in World Models}

\author{
    Dingyi Rong\textsuperscript{\rm 1,\rm 2},
    Yue Shi\textsuperscript{\rm 2$\dagger$},
    Chaofan Ma\textsuperscript{\rm 1$\ddagger$},
    Jiezhang Cao\textsuperscript{\rm 1},
    Zongrui Wang\textsuperscript{\rm 1,\rm 2},\\
    Zeyu Zhang\textsuperscript{\rm 1,\rm 2},
    Yao Mu\textsuperscript{\rm 1,\rm 2},
    Guangtao Zhai\textsuperscript{\rm 1,\rm 2$\dagger$},
    Ning Liu\textsuperscript{\rm 1$\dagger$}
}
\affiliations{
    \textsuperscript{\rm 1}Shanghai Jiao Tong University\\
    \textsuperscript{\rm 2}Shanghai Artificial Intelligence Laboratory
}

\begin{document}

\maketitle
{\renewcommand{\thefootnote}{\fnsymbol{footnote}}%
\footnotetext[2]{Corresponding authors.}%
\footnotetext[3]{Project lead.}}

\begin{abstract}
Large-scale manipulation data is essential for robot learning, yet collecting robot demonstrations remains expensive and difficult to scale. Meanwhile, abundant egocentric human manipulation videos provide rich behavioral experiences, but transferring them across embodiments remains challenging due to differences between human hands and robotic end-effectors. Recent advances in video world models offer a promising pathway to synthesize robot-centric manipulation videos from human observations, while their cross-embodiment transfer capability remains largely unexplored.
Therefore, we introduce H2R-Bench, a benchmark for evaluating cross-embodiment human-to-robot manipulation video generation, where models transform egocentric human demonstrations into robot manipulation videos under specified embodiments.
Each benchmark instance contains a human demonstration video, target embodiment constraints, and source-grounded annotations covering task goals, action events, functional contacts, and object responses.
H2R-Bench evaluates generated videos through five dimensions, including goal-state completion, action-event completion, functional contact transfer, embodiment correctness, and general video quality. We benchmark eleven state-of-the-art video generation models across six manipulation families and two robot embodiments. Our evaluation reveals that current video world models remain limited in human-to-robot manipulation transfer: even leading models often fail in embodiment consistency, functional interaction, and task execution. H2R-Bench provides a systematic diagnostic framework for evaluating whether video world models can bridge the human-to-robot embodiment gap and convert human manipulation observations into robot-centric training resources. Project page: \url{https://rongdingyi.github.io/H2R-Bench/}

\end{abstract}

\begin{figure}[t]
\centering
\includegraphics[width=0.98\columnwidth]{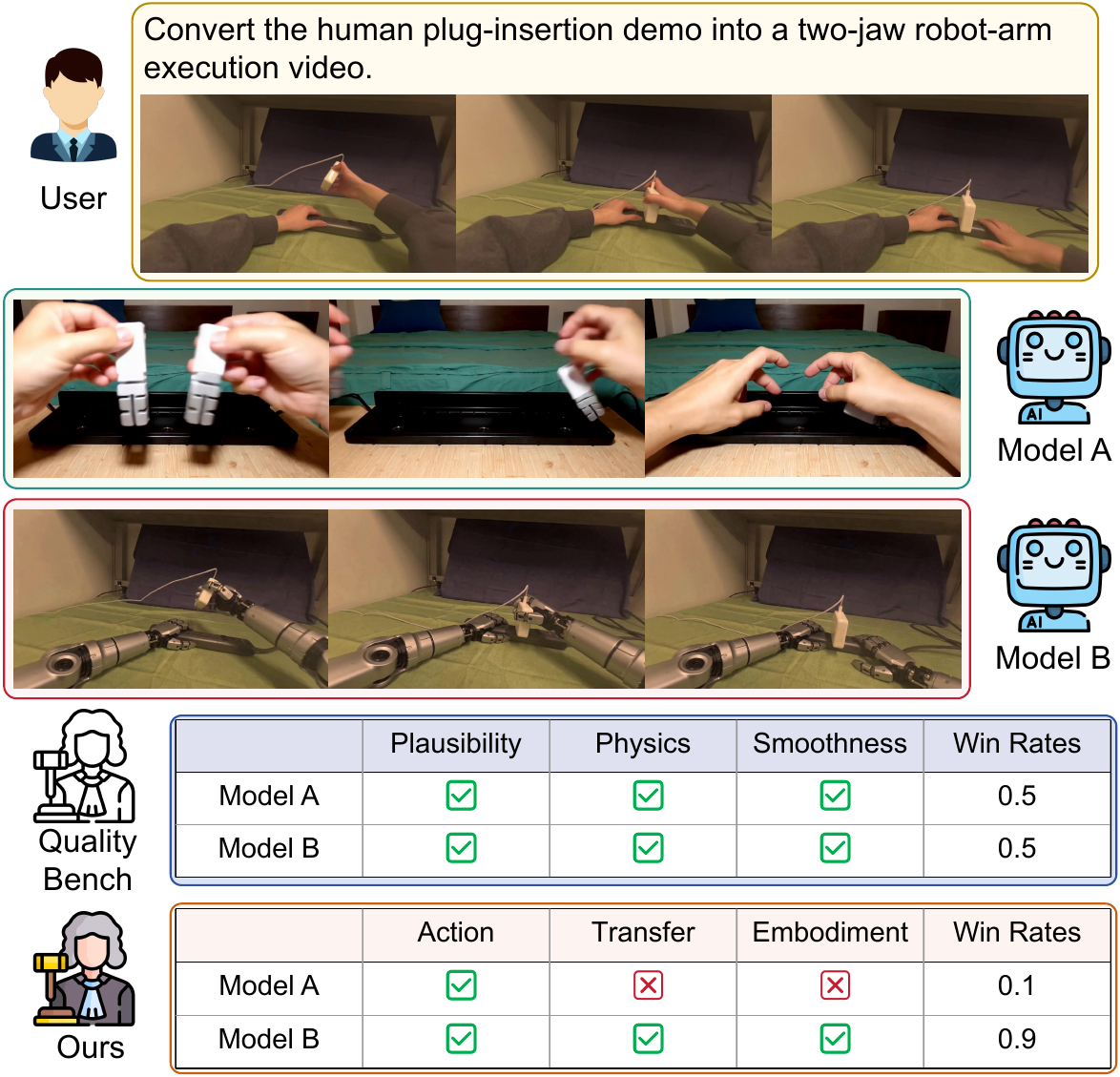}
\caption{Comparison between existing evaluation and H2R-Bench. Given a human demonstration and a target embodiment instruction, video world models generate robot manipulation videos. Existing video benchmarks such as WorldModelBench \cite{li2026worldmodelbench} assess overall video plausibility, rates both videos highly, whereas our H2R-Bench diagnoses transfer through goal, action, contact, and embodiment.}
\vspace{-0.4cm}
\label{fig:h2r_motivation}
\end{figure}

\section{Introduction}
Recent progress in robot learning has shown the importance of large-scale demonstrations across diverse tasks, objects, and environments \cite{brohan2022rt, zitkovich2023rt, o2024open, black2024pi_0, zhang2026r3dp}. However, collecting robot manipulation videos remains expensive, requiring specific hardware, teleoperation interfaces, calibrated cameras, safety constraints, and repeated physical execution. In contrast, egocentric human videos are abundant and naturally capture everyday object affordances, hand-object contacts, manipulation intent, and physical state changes \cite{damen2022rescaling, grauman2022ego4d, hoque2025egodex, ma2026robot, ma2026reason}. This asymmetry has motivated human-to-robot and cross-embodiment generation studies, which explore video generation as a bridge between abundant human demonstrations and scarce robot demonstrations \cite{xie2026human2robot, song2025mitty, zhang2026qwen}. In this view, a generated robot video serves as an intermediate representation of a target robot execution conditioned on a human demonstration. The key challenge is not merely generating a visually plausible robot video, but preserving the manipulation evidence in the source demonstration while adapting the execution to a different embodiment. As illustrated in Figure~\ref{fig:h2r_motivation}, evaluating such videos requires criteria beyond frame-level realism and temporal coherence. A valid human-to-robot video should preserve the task intent, required actions, and functional interactions implied by the source demonstration.

\begin{table}[t]
\centering
\small
\setlength{\tabcolsep}{2.3pt}
\renewcommand{\arraystretch}{1.03}
\begin{tabular}{@{}lcccccccc@{}}
\toprule
\multirow{2}{*}{Benchmark} & \multicolumn{3}{c}{Settings} & \multicolumn{5}{c}{Evaluation} \\
\cmidrule(lr){2-4}\cmidrule(lr){5-9}
& I2V & RV & H2R & VQ & Goal & Action & Cont. & Emb. \\
\midrule
VBench & $\times$ & $\times$ & $\times$ & $\checkmark$ & $\times$ & $\times$ & $\times$ & $\times$ \\
WorldModelBench & $\checkmark$ & $\triangle$ & $\times$ & $\checkmark$ & $\triangle$ & $\triangle$ & $\times$ & $\times$ \\
RBench & $\checkmark$ & $\checkmark$ & $\times$ & $\checkmark$ & $\checkmark$ & $\checkmark$ & $\times$ & $\triangle$ \\
RoboWM-Bench & $\checkmark$ & $\checkmark$ & $\times$ & $\triangle$ & $\checkmark$ & $\checkmark$ & $\times$ & $\times$ \\
RoboTrustBench & $\checkmark$ & $\checkmark$ & $\times$ & $\checkmark$ & $\checkmark$ & $\checkmark$ & $\times$ & $\triangle$ \\
\midrule
\textbf{H2R-Bench (ours)} & $\checkmark$ & $\checkmark$ & $\checkmark$ & $\checkmark$ & $\checkmark$ & $\checkmark$ & $\checkmark$ & $\checkmark$ \\
\bottomrule
\end{tabular}
\caption{Comparison of H2R-Bench and existing video-generation benchmarks across evaluation capabilities. ``I2V'', ``RV'', and ``H2R'' denote image-to-video generation, robot-video evaluation, and human-to-robot transfer, respectively. 
Evaluation dimensions include visual quality, goal completion, action completion, functional contact transfer, and embodiment consistency. 
$\checkmark$, $\triangle$, and $\times$ indicate full, partial, and no support, respectively.}
\label{tab:benchmark_comparison}
\end{table}

As summarized in Table~\ref{tab:benchmark_comparison}, existing evaluation protocols only partially capture this ability. General video-generation benchmarks mainly measure visual fidelity, temporal consistency, text-video alignment, and low-level motion quality \cite{huang2024vbench, liu2024evalcrafter, sun2025t2v, ji2024t2vbench}. Recent world-model and robotics-oriented benchmarks further examine physical plausibility, action completeness, robot structure, executability, or trustworthiness in manipulation videos \cite{li2026worldmodelbench, bansal2025videophy, han2026oscbench, deng2026rethinking, jiang2026robowm, li2026robotrustbench}. However, these benchmarks typically condition generation on text, images, initial robot states, or robot-centric scenarios and do not evaluate cross-embodiment manipulation transfer from human demonstrations to robot embodiments. As a result, existing benchmarks cannot diagnose H2R-specific failures, such as incorrect embodiment replacement or interactions that are inconsistent with the source demonstration.

We introduce H2R-Bench, a benchmark for evaluating source-conditioned human-to-robot video transfer. Unlike existing evaluations that focus on generating plausible robot videos from language prompts or robot-centric states, H2R-Bench investigates whether models can transform an egocentric human demonstration into a robot manipulation video under a specified embodiment. Given a source human video, H2R-Bench evaluates whether the generated video preserves the demonstrated task goal, required action events, functional interactions, and object-state changes, while realizing the execution with a robot morphology and end-effector consistent with the target embodiment. H2R-Bench covers six task families defined by the physical state change that determines success: rigid-object rearrangement, mechanism actuation, insertion and assembly, deformable-object configuration, bulk-material transfer, and surface or material transformation. Each demonstration is paired with a human-verified structured task specification, initialized by a multimodal language model and reviewed against the source video, describing the intended goal, required action events, manipulated entities, relevant state transitions, and source-side contact evidence, thereby defining what must be preserved without prescribing a specific robot trajectory.

H2R-Bench scores five complementary dimensions: goal completion, action-event completion, functional contact transfer, embodiment correctness, and task-agnostic video quality. Their weighted aggregate H2RCore emphasizes contact and embodiment while retaining a smaller contribution from video quality. We evaluate 11 representative video generation models, including five proprietary models \cite{seedance2026seedance,wan2025wan,team2025kling,google2025veo31,xai2026grokimagine} and six open-source models \cite{wan2025wan,hacohen2026ltx,wu2025hunyuanvideo,li2026skyreels,team2025longcat,song2025mitty}. Although M5 occupies a narrow range of 0.73--0.81, H2RCore spans 30.0--84.6 and has weak rank association with M5 ($\rho=0.14$). Three human raters validate the transfer metrics, with within-scene Spearman $\rho=0.883$ and aggregate Pearson $r=0.930$ between human and automated transfer subscores.

Our contributions are summarized as follows:
\begin{itemize}
    \item We introduce a source-relative evaluation setting for human-to-robot video transfer, requiring generated videos to preserve the task goal and functional interaction of a specific human demonstration while replacing the human actor with a specified robot embodiment.
    
    \item We introduce H2R-Bench, spanning six manipulation families and two target embodiments, with structured annotations of task goals, action events, object-state evolution, and functional contact.
    
    \item We develop a transfer-aware evaluation protocol that separately measures goal-state completion, action-event completion, functional contact transfer, embodiment correctness, and task-agnostic video quality, and use it to reveal a persistent mismatch between visual quality and valid robot transfer in current video generation models.
\end{itemize}

\section{Related Work}

\begin{figure*}[t]
\centering
\includegraphics[width=0.98\textwidth]{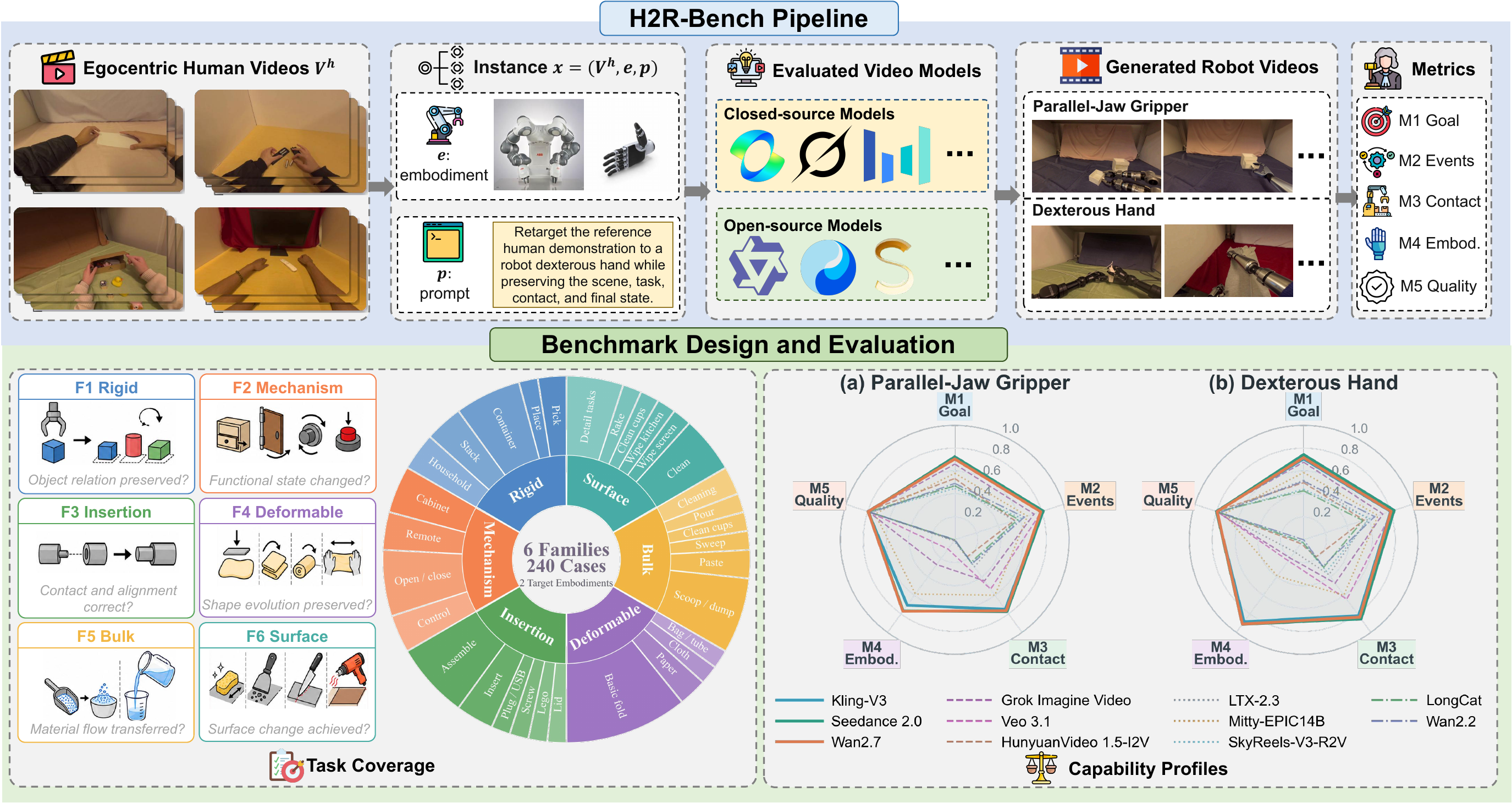}
\caption{Overview of H2R-Bench. The benchmark curates egocentric human manipulation demonstrations, conditions video generators on each model's supported source interface and target embodiment, and evaluates the resulting robot videos with transfer-aware metrics. The lower panels summarize task coverage and model capability profiles across both target embodiments.}
\label{fig:h2r_overview}
\end{figure*}

\subsection{Video Generation and World Models for Robotics}

\noindent Video generation has advanced toward controllable models with stronger temporal coherence, visual consistency, and interaction modeling~\cite{blattmann2023align,bar2024lumiere,kong2024hunyuanvideo,wan2025wan,wang2026rein3d}. Recent studies further view these generators as predictive world models of object dynamics, state transitions, and interaction outcomes~\cite{brooks2024video,bruce2024genie,wang2026deepgen,wang2026unireason}. This perspective is relevant to robotics, where generated videos can represent future observations and provide scalable data beyond costly robot demonstrations.

\subsection{Human-to-Robot Video Transfer}

Egocentric human videos provide scalable supervision because they expose task goals, object affordances, hand--object interactions, and state changes~\cite{grauman2022ego4d,damen2022rescaling,hoque2025egodex,ma2026robot}. Prior work transfers this information through cross-embodiment visual representations and reward learning~\cite{nair2022r3m,ma2022vip,ma2023liv}, or through affordance, trajectory, and latent-action abstractions for downstream manipulation~\cite{bahl2023affordances,bharadhwaj2024track2act,chen2025vidbot,ye2025latent,chen2025moto,li2025scalable,zheng2026egoscale}.
A complementary line directly constructs robot-oriented observations or data through visual translation, editing, rendering, retargeting, and generative modeling~\cite{smith2019avid,lepert2025masquerade,li2025h2r,li2025mimicdreamer}. Mitty and H2R-Grounder generate robot videos from human interaction videos~\cite{song2025mitty,ci2025h2r}, while Human2Robot learns action transfer from paired human--robot demonstrations~\cite{xie2026human2robot}. These studies motivate evaluating how faithfully video generators transfer manipulation evidence across embodiments.

\subsection{Evaluation of Video World Models for Robotics}

Video-generation benchmarks assess perceptual quality, temporal consistency, text--video alignment, motion, and compositional correctness~\cite{huang2024vbench,liu2024evalcrafter,sun2025t2v,ji2024t2vbench}. Recent world-model and robotics benchmarks additionally examine physical commonsense, object-state changes, robot appearance, action completeness, executability, and trustworthiness~\cite{bansal2025videophy,li2026worldmodelbench,han2026oscbench,deng2026rethinking,jiang2026robowm,li2026robotrustbench,xia2026roboprocessbench}.
They nevertheless evaluate videos against text, initial visual conditions, or robot-centric tasks. Human-to-robot transfer instead requires a robot video to remain faithful to a particular human demonstration while realizing a requested embodiment. This source-relative setting jointly requires task goal, required action events, functional contact, object response, and target-embodiment realization; existing benchmarks evaluate only subsets of these properties.

\section{H2R-Bench}

\subsection{Task Formulation}

H2R-Bench evaluates video transfer from a human demonstration to a specified robot embodiment. Given an egocentric human manipulation video and a target embodiment, the model generates a robot video of the same task. The source video is treated as evidence of what happened, not simply as a visual style reference. A successful output should retain the task goal, the actions needed to reach it, and the visible interaction between the actor and the objects. We do not require pose or trajectory imitation: a parallel-jaw gripper and a dexterous hand may solve the task differently, as long as the new strategy remains compatible with the requested morphology and the source task.

Each case is
\begin{equation}
x_i=(V_i^h,e_i,p_i),
\end{equation}
where $V_i^h$ is the human source video, $e_i$ is the target embodiment, and $p_i$ is the generation prompt. The prompt names the task goal and target embodiment and asks the model to preserve the source scene, objects, and interaction. It does not contain evaluation weights, evidence-frame indices, or per-metric checks. Those fields are kept in a separate annotation used only for scoring. Model $m$ receives the visual input supported by its interface, denoted by $\mathcal{Z}_{i,m}^h$: the full source video when video conditioning is available, or an ordered subset of source frames otherwise. It generates
\begin{equation}
V_{i,m}^r=G_m\bigl(\mathcal{Z}_{i,m}^h,p_i\bigr).
\end{equation}
Models without visual source conditioning are outside this source-conditioned setting.

\subsection{Benchmark Construction}
\label{sec:benchmark_construction}

H2R-Bench uses 120 egocentric clips from the EgoDex test split~\cite{hoque2025egodex} as source videos. We select clips with visible task-relevant entities, interactions, and state changes, and form 240 transfer cases by pairing each of 120 source videos with two target embodiments: a parallel-jaw gripper and a dexterous hand. The sources are evenly distributed across six manipulation families: rigid-object transport or rearrangement (F1), articulated-mechanism actuation (F2), insertion or attachment (F3), deformable-object configuration (F4), bulk-material transfer or mixing (F5), and surface or material change (F6).

Qwen3.7-Plus produces an annotation for each 5s source clip from the video. It records the initial and final states, relevant objects and tools, required action events, and source-side contact evidence. For each target embodiment, the annotation also specifies functional contact regions, manipulation modes, expected object responses, and supporting or bimanual roles when applicable. All benchmark annotations are manually verified against their source videos.

\subsection{Native-Interface Generation Protocol}

All models use the same source cases, target embodiments, task specifications, and evaluation criteria. We evaluate each model with its strongest publicly documented source-conditioning interface. Video-conditioned models receive the full source clip. Image/frame-conditioned models receive ordered source frames up to the interface limit, with the temporal endpoints retained whenever the interface allows more than one image. The wording of the task and embodiment prompt is kept aligned across models.

No target-robot reference image is used in the main setting. The target is specified in text, which tests whether the model can replace the human actor while preserving the source interaction. We study robot reference images in a separate ablation below. Generated videos remain in their native resolution, duration, and frame rate. At evaluation time, we uniformly sample a fixed number of frames for each metric, so every model is judged with the same evidence budget.

\subsection{Transfer-Aware Evaluation}

H2R-Bench reports five complementary scores for each generated video. M1--M4 measure whether the source manipulation has been transferred correctly, covering the intended final state, required action events, functional robot--object contact, and the requested robot embodiment. M5 separately measures task-agnostic video quality. For M1--M4, three MLLM judges independently assess the prescribed visual evidence using a shared 0--4 rubric, where 0 indicates absent or contradictory evidence and 4 indicates clear and complete satisfaction. Judge scores are normalized to $[0,1]$ and averaged.

\paragraph{M1: Goal-State Completion.} Each case defines a set of weighted predicates describing the final state required for task success. Depending on the task, these predicates may specify a spatial or containment relation, a mechanism state, an attachment, a deformation, a material distribution, or a surface change. Judges receive 25 uniformly sampled frames: the sequence provides context, while the final frames determine whether each predicate has been satisfied. The weighted average of the normalized predicate scores gives $S_{\mathrm{goal}}$.

\paragraph{M2: Action-Event Completion.} Each case also specifies the action events required to complete the task, together with their relative weights. Using the 25 sampled frames, judges determine how clearly each event is carried out. Typical events include grasping, inserting, releasing, pouring, wiping, and folding. The resulting weighted average defines $S_{\mathrm{action}}$. This metric concerns whether the required operations occur, rather than whether the robot reproduces the human pose, trajectory, or timing.

\paragraph{M3: Functional Contact Transfer.} M3 examines whether the generated robot establishes an interaction that is functionally equivalent to the one demonstrated by the human. Judges compare 25 source frames with 25 generated frames, together with a contact specification derived from the source video. The assessment considers the contacted functional region, whether contact is visibly established, the manipulation mode, whether the object response follows from that contact, and whether the interaction is feasible for the requested embodiment. The robot may use a different grasp or trajectory, provided that the contact serves the same manipulation function. Non-applicable dimensions are omitted when computing $S_{\mathrm{contact}}$.

\paragraph{M4: Embodiment Correctness.} M4 assesses whether the generated actor consistently matches the requested robot embodiment. The rubric separates robot presence from morphology: judges examine whether a robot is visible, whether human hands remain, whether the requested embodiment class is used, whether the end-effector type is correct, and whether the robot structure remains consistent over time. This distinction is important because a visible robot arm may still carry an incorrect gripper or hand. If no robot is present, or a human performs the main manipulation, $S_{\mathrm{emb}}$ is set to zero.

\paragraph{M5: Video Quality.} Unlike M1--M4, M5 is computed without reference to the task annotation. It averages four normalized components: imaging quality, estimated framewise with MUSIQ~\cite{ke2021musiq}; aesthetic quality, obtained from the LAION linear predictor over normalized CLIP ViT-L/14 features~\cite{radford2021clip,laion2022aesthetic}; temporal stability, measured from adjacent-frame differences; and motion smoothness, measured by the reconstruction error of AMT-S midpoint interpolation~\cite{li2023amt}. Their arithmetic mean defines $S_{\mathrm{video}}$.

\begin{table*}[!ht]
\centering
\small
\setlength{\tabcolsep}{2.6pt}
\begin{tabular}{lrrrrr@{\hspace{3.5pt}}c@{\hspace{5pt}}rrrrr@{\hspace{3.5pt}}c}
\toprule
\multicolumn{1}{c}{\multirow{3}{*}{Model}} & \multicolumn{6}{c}{Parallel-Jaw Gripper} & \multicolumn{6}{c}{Dexterous Hand} \\
\cmidrule(lr){2-7}\cmidrule(lr){8-13}
& \multicolumn{5}{c}{Component Metrics} & \multicolumn{1}{c}{Aggregate}
& \multicolumn{5}{c}{Component Metrics} & \multicolumn{1}{c}{Aggregate} \\
\cmidrule(lr){2-6}\cmidrule(r){7-7}\cmidrule(lr){8-12}\cmidrule(r){13-13}
 & \shortstack{Goal\\Comp.} & \shortstack{Action\\Comp.} & \shortstack{Contact\\Transfer} & \shortstack{Embod.\\Correct.} & \shortstack{Video\\Quality} & \shortstack{\textbf{H2R}\\\textbf{Core}} & \shortstack{Goal\\Comp.} & \shortstack{Action\\Comp.} & \shortstack{Contact\\Transfer} & \shortstack{Embod.\\Correct.} & \shortstack{Video\\Quality} & \shortstack{\textbf{H2R}\\\textbf{Core}} \\
\midrule
\rowcolor{h2rvideocond}\multicolumn{13}{l}{\textit{Video-conditioned generation}} \\
\rowcolor{h2rvideocond}Seedance 2.0 & \textbf{0.725} & \textbf{0.813} & \textbf{0.776} & \underline{0.768} & 0.793 & \textbf{77.3} & \textbf{0.744} & \textbf{0.832} & \textbf{0.855} & \textbf{0.911} & 0.799 & \textbf{84.6} \\
\rowcolor{h2rvideocond}Wan2.7 & 0.706 & 0.791 & \underline{0.766} & \textbf{0.772} & 0.796 & \underline{76.5} & \underline{0.718} & 0.804 & \underline{0.835} & \underline{0.910} & 0.795 & \underline{83.1} \\
\rowcolor{h2rvideocond}Kling-V3 & 0.710 & \underline{0.807} & 0.751 & 0.707 & \underline{0.798} & 74.5 & 0.707 & 0.800 & 0.819 & 0.885 & 0.802 & 81.7 \\
\midrule
\rowcolor{h2rframecond}\multicolumn{13}{l}{\textit{Frame-conditioned generation}} \\
\rowcolor{h2rframecond}Mitty-EPIC14B & 0.581 & 0.668 & 0.598 & 0.585 & 0.732 & 61.5 & 0.587 & 0.684 & 0.598 & 0.392 & 0.732 & 56.1 \\
\rowcolor{h2rframecond}Veo 3.1 & \textbf{0.725} & 0.797 & 0.533 & 0.100 & 0.783 & 49.6 & 0.715 & \underline{0.816} & 0.642 & 0.227 & 0.793 & 57.0 \\
\rowcolor{h2rframecond}Grok Imagine Video & 0.661 & 0.729 & 0.443 & 0.268 & 0.792 & 50.1 & 0.678 & 0.729 & 0.469 & 0.198 & \underline{0.804} & 49.2 \\
\rowcolor{h2rframecond}LTX-2.3 & 0.473 & 0.545 & 0.292 & 0.012 & 0.773 & 32.1 & 0.520 & 0.592 & 0.377 & 0.132 & 0.780 & 39.8 \\
\rowcolor{h2rframecond}SkyReels-V3-R2V & 0.448 & 0.610 & 0.256 & 0.004 & 0.787 & 31.5 & 0.441 & 0.607 & 0.341 & 0.026 & 0.789 & 34.6 \\
\rowcolor{h2rframecond}Wan2.2 & 0.492 & 0.639 & 0.258 & 0.000 & 0.769 & 32.4 & 0.512 & 0.653 & 0.286 & 0.000 & 0.766 & 33.7 \\
\rowcolor{h2rframecond}LongCat & 0.472 & 0.583 & 0.243 & 0.000 & 0.790 & 31.0 & 0.428 & 0.545 & 0.298 & 0.020 & 0.793 & 32.0 \\
\rowcolor{h2rframecond}HunyuanVideo 1.5-I2V & 0.535 & 0.549 & 0.184 & 0.005 & \textbf{0.806} & 30.0 & 0.499 & 0.555 & 0.185 & 0.041 & \textbf{0.808} & 30.7 \\
\bottomrule
\end{tabular}
\caption{Main-evaluation results by target embodiment. M1--M5 measure goal completion, action completion, contact transfer, embodiment correctness, and Video Quality; H2RCore aggregates all five metrics on a 0--100 scale. Rows are ordered by the sum of the two H2RCore scores within each conditioning group. \textbf{Bold} and \underline{underlined} entries indicate the best and second-best result in each column.}
\label{tab:main_results_by_embodiment}
\end{table*}

The primary benchmark score combines the five normalized components defined above:
\begin{equation}
\begin{aligned}
\mathrm{H2RCore}=100\bigl(&0.15S_{\mathrm{goal}}+0.15S_{\mathrm{action}}\\
&+0.30S_{\mathrm{contact}}+0.30S_{\mathrm{emb}}\\
&+0.10S_{\mathrm{video}}\bigr).
\end{aligned}
\end{equation}
For M1--M4, the three judge scores are averaged for each video. Dataset-level component scores are then computed over videos, and H2RCore is obtained from the component means. We assign 0.30 each to contact and embodiment, which account for 60\% of the score: valid transfer requires both a functionally supported interaction and the requested robot morphology, and equal weights avoid privileging one requirement over the other. Goal and action completion each receive 0.15. These terms preserve the source task, but neither is sufficient for transfer because a human-led or wrong-embodiment video may still show the correct outcome and actions. Video quality receives 0.10, so visual polish contributes to the score without compensating substantially for failures in contact or embodiment. Supplementary~\ref{sec:statistical_reporting} reports sensitivity to alternative weight choices.

\section{Experiments}
\label{sec:experiments}

We evaluate 11 current video generators on the 240 transfer cases using their native source-conditioning interfaces. Beyond the leaderboard, we examine how the target embodiment changes performance, whether a target-robot reference image helps, and whether generic video quality tracks transfer quality. Human scores and qualitative examples provide an additional check on the automated metrics.

\subsection{Experimental Setup}

\paragraph{Benchmark Instances.}
The main evaluation contains 240 transfer cases constructed from 120 human source videos. Each source is paired with both target embodiments, so the two conditions are evaluated on matched demonstrations. The sources are evenly distributed across the six manipulation families, with 20 sources per family.

\paragraph{Models.}
We evaluate five proprietary models (Seedance 2.0 \cite{seedance2026seedance}, Wan2.7 \cite{wan2025wan}, Kling-V3 \cite{team2025kling}, Veo 3.1 \cite{google2025veo31}, Grok Imagine Video \cite{xai2026grokimagine}), and six open-source models (Wan2.2 \cite{wan2025wan}, LTX-2.3 \cite{hacohen2026ltx}, HunyuanVideo 1.5-I2V \cite{wu2025hunyuanvideo}, SkyReels-V3-R2V \cite{li2026skyreels}, LongCat \cite{team2025longcat}, Mitty-EPIC14B \cite{song2025mitty}). Together they cover full-video and frame-conditioned generation models. 

\paragraph{Judges and Metrics.}
Gemini 3.5 Flash, Qwen3.7-Plus, and GPT-5.4 independently score M1--M4, and we average the three judgments. The metrics use 25 uniformly sampled frames from each video. M3 receives 25 frames from both the source and generated videos. M5 averages normalized MUSIQ imaging quality, CLIP-based aesthetic quality, adjacent-frame temporal stability, and AMT-S interpolation consistency. H2RCore determines the transfer ranking. M1--M5 remain visible in the tables to distinguish task, contact, and embodiment failures.

\paragraph{Native-Interface Protocol.}
Seedance 2.0, Wan2.7, and Kling-V3 receive the source video. The other models receive ordered source frames within their interface limits. All models use the same task and embodiment information.

\subsection{Main Results}

Table~\ref{tab:main_results_by_embodiment} compares all 11 models under both target embodiments. The three video-conditioned systems occupy the top positions. Seedance 2.0 ranks first with H2RCore scores of 77.3 for the Parallel-Jaw Gripper and 84.6 for the Dexterous Hand, followed by Wan2.7 (76.5 and 83.1) and Kling-V3 (74.5 and 81.7). Among these models, goal and action scores are similar; most of the separation comes from contact transfer and embodiment correctness.

The metric breakdown explains why broad task recognition is not enough. Veo 3.1 attains the best Parallel-Jaw Gripper M1 score (0.725) and a Dexterous Hand M2 score of 0.816, yet its M4 scores are only 0.100 and 0.227. HunyuanVideo 1.5-I2V achieves the highest M5 for both targets (0.806 and 0.808), but its contact score stays near 0.185 and its embodiment score near zero. Both models often preserve a recognizable task or polished appearance without completing the requested robot transfer.

Within the video-conditioned subgroup, Seedance's lead over Wan2.7 is small but consistent. The paired H2RCore difference is $+0.79$ with a 95\% confidence interval of $[+0.08,+1.48]$ for the gripper and $+1.52$ with $[+0.92,+2.16]$ for the hand. Comparisons with Kling-V3 are farther from zero (Supplementary Table~\ref{tab:leading_pairwise_ci}).

\begin{table}[t]
\centering
\small
\setlength{\tabcolsep}{3pt}
\renewcommand{\arraystretch}{1.08}
\begin{tabular}{lcc}
\toprule
Metric & \shortstack{Mean change\\(Hand $-$ Gripper)} & \shortstack{Hand higher\\(models)} \\
\midrule
Goal-State Completion (M1) & $+0.002$ & 6/11 \\
Action-Event Completion (M2) & $+0.008$ & 8/11 \\
Functional Contact Transfer (M3) & $+0.055$ & 11/11 \\
Embodiment Correctness (M4) & $+0.047$ & 8/11 \\
Video Quality (M5) & $+0.004$ & 8/11 \\
H2RCore (0--100) & $+3.3$ & 9/11 \\
\bottomrule
\end{tabular}
\caption{Effect of target embodiment across all 11 models. Mean change reports the average score difference between the Dexterous Hand and Parallel-Jaw Gripper. ``Hand higher'' reports the number of models with a positive difference.}
\label{tab:embodiment_shift}
\end{table}

\begin{table}[t]
\centering
\footnotesize
\setlength{\tabcolsep}{0pt}
\renewcommand{\arraystretch}{1.08}
\begin{tabular*}{\columnwidth}{@{\extracolsep{\fill}}llrrrrrr@{}}
\toprule
Model & Ref. & Goal & Action & Contact & Embod. & Quality & Core \\
\midrule
\multicolumn{8}{l}{\textit{Parallel-Jaw Gripper}} \\
\multirow{2}{*}{Kling-V3} & No & 0.710 & 0.807 & 0.751 & 0.707 & 0.798 & 74.5 \\
 & Yes & 0.673 & 0.760 & 0.577 & 0.479 & 0.779 & 61.0 \\
\multirow{2}{*}{Seedance 2.0} & No & 0.725 & 0.813 & 0.776 & 0.768 & 0.793 & 77.3 \\
 & Yes & 0.711 & 0.794 & 0.674 & 0.595 & 0.784 & 68.5 \\
\multirow{2}{*}{Wan2.7} & No & 0.706 & 0.791 & 0.766 & 0.772 & 0.796 & 76.5 \\
 & Yes & 0.714 & 0.811 & 0.871 & 0.875 & 0.781 & 83.1 \\
\midrule
\multicolumn{8}{l}{\textit{Dexterous Hand}} \\
\multirow{2}{*}{Kling-V3} & No & 0.707 & 0.800 & 0.819 & 0.885 & 0.802 & 81.7 \\
 & Yes & 0.629 & 0.734 & 0.772 & 0.728 & 0.799 & 73.4 \\
\multirow{2}{*}{Seedance 2.0} & No & 0.744 & 0.832 & 0.855 & 0.911 & 0.799 & 84.6 \\
 & Yes & 0.736 & 0.824 & 0.876 & 0.753 & 0.794 & 80.2 \\
\multirow{2}{*}{Wan2.7} & No & 0.718 & 0.804 & 0.835 & 0.910 & 0.795 & 83.1 \\
 & Yes & 0.712 & 0.819 & 0.905 & 0.873 & 0.789 & 84.2 \\
\bottomrule
\end{tabular*}
\caption{Effects of target-robot reference images for three video-conditioned models. ``No'' and ``Yes'' indicate generation without and with a target-robot reference image.}
\label{tab:robot_reference_ablation}
\end{table}

\begin{figure}[!ht]
\centering
\includegraphics[width=0.98\columnwidth]{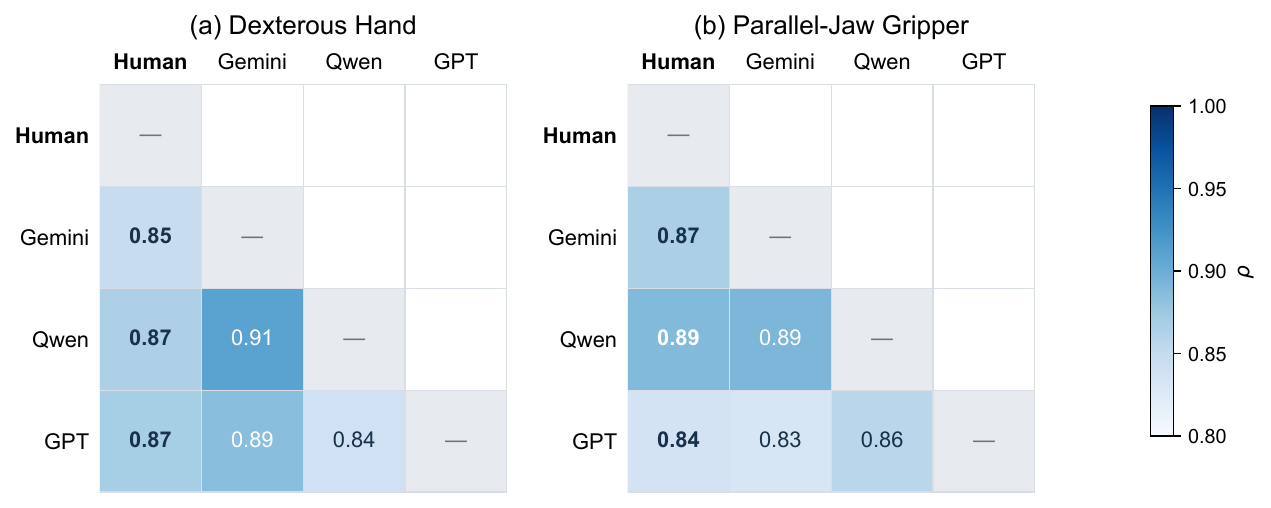}
\caption{Agreement between human and MLLM evaluators. Human raters and MLLM judges rank generated videos based on the transfer score aggregated from M1–M4. MLLM-based evaluation closely aligns with human judgments, with Spearman correlations above 0.8 across evaluators.}
\label{fig:human_mllm_spearman}
\end{figure}

\begin{figure}[!ht]
\centering
\includegraphics[width=0.98\columnwidth]{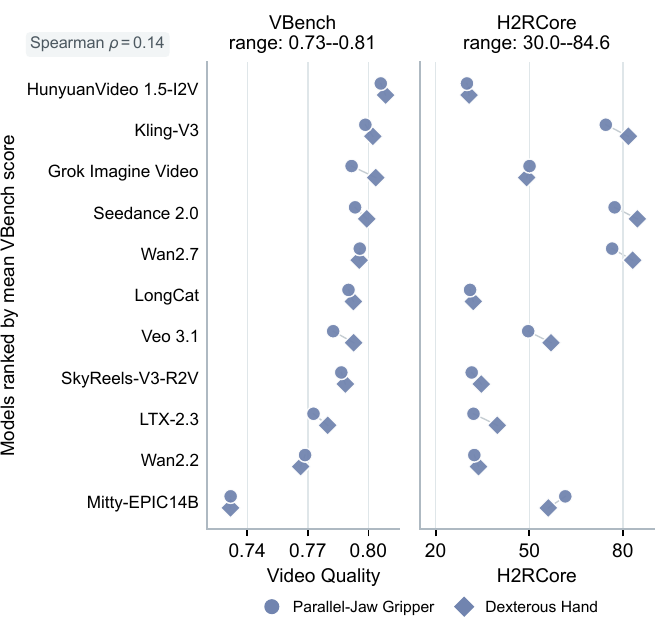}
\caption{VBench Video Quality versus H2RCore. Unlike VBench, H2RCore better differentiates the models.}
\label{fig:m5_vs_h2rcore}
\end{figure}

\begin{figure*}[!ht]
\centering
\includegraphics[width=0.99\textwidth]{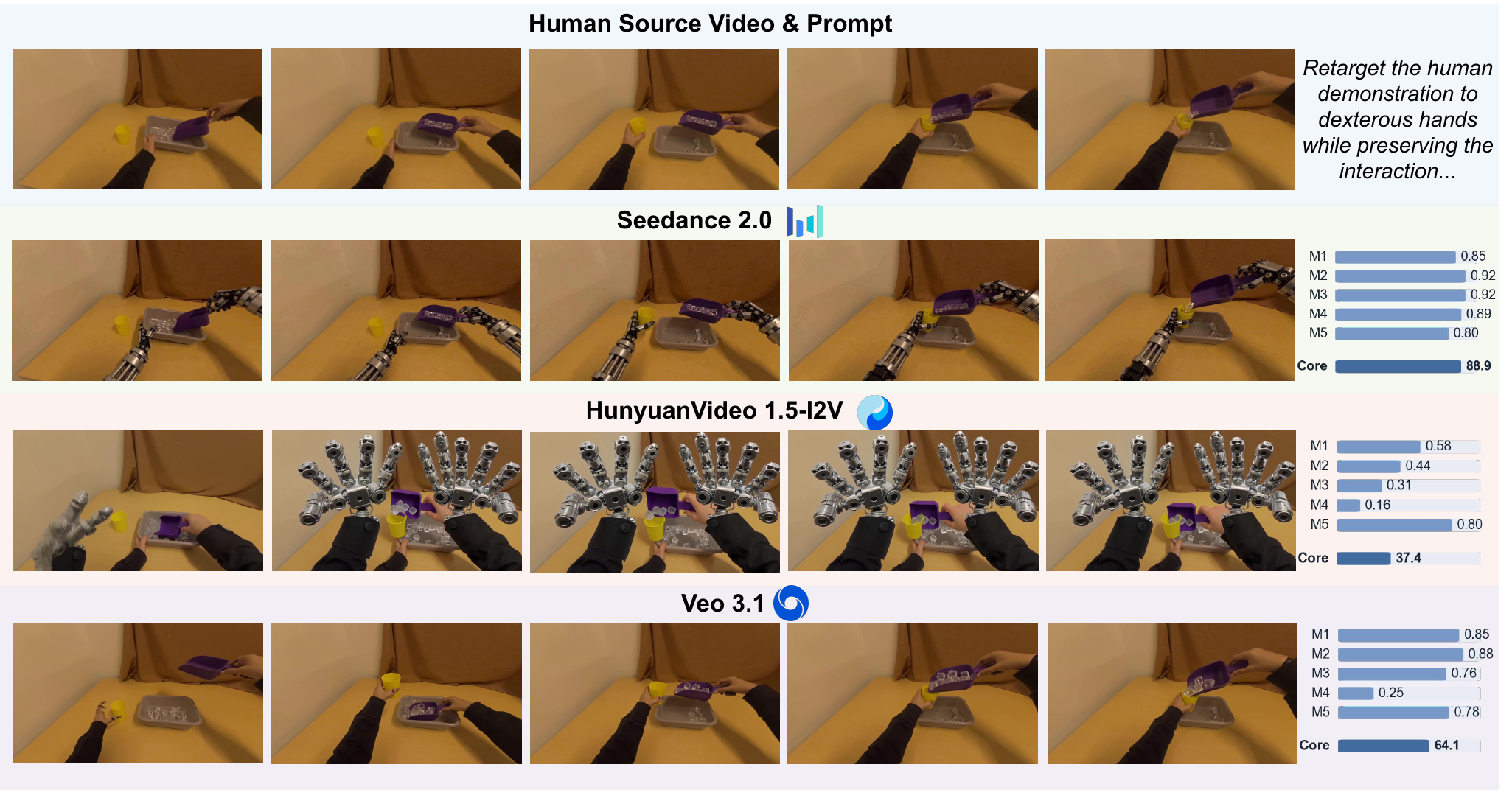}
\caption{Qualitative H2R transfer results with the shared source video and prompt. The unscored top row is the human source; the generated rows show representative stages of each output. Metric strips report per-video M1--M5 and H2RCore.}
\label{fig:qualitative_metrics}
\end{figure*}

\subsubsection{Human Evaluation.}
Three human raters independently score M1--M4 for all 11 models under both target embodiments on five sampled source scenes per task family, totaling 660 generated videos. Figure~\ref{fig:human_mllm_spearman} compares within-scene rankings computed from the corresponding transfer subscore. The macro-average Human--MLLM correlation is $\rho=0.883$, showing that the automated evaluation largely preserves human model preferences within a shared task. M5 relies on established metrics that have been independently validated in prior work, and is therefore excluded from the human-agreement analysis.
\subsection{Embodiment Effects}

\paragraph{Gripper vs. Dexterous Hand.}
Table~\ref{tab:embodiment_shift} compares the same 120 sources under the two target embodiments. Nine of the 11 models score higher with the Dexterous Hand, with an average H2RCore gain of 3.3 points. Goal, action, and Video Quality change little, whereas contact transfer rises by $0.055$ for all 11 models and embodiment correctness by $0.047$ across eight. The effect is especially clear for the three video-conditioned models: the Dexterous Hand raises H2RCore by 7.3 points for Seedance, 6.6 for Wan2.7, and 7.2 for Kling-V3. Its morphology is closer to the human actor in the source video, which is consistent with better preservation of contact during actor replacement. This advantage is not universal, however: Grok Imagine Video and Mitty-EPIC14B both score lower with the hand.

\paragraph{Robot Reference Image.}

Table~\ref{tab:robot_reference_ablation} shows that a target-robot reference image does not provide a uniform benefit. Wan2.7 gains the most. For the gripper, contact transfer rises from 0.766 to 0.871 and embodiment correctness from 0.772 to 0.875, increasing H2RCore from 76.5 to 83.1. Its Dexterous Hand score improves more modestly, from 83.1 to 84.2. Kling-V3 and Seedance move in the opposite direction. Kling loses 13.5 H2RCore points for the gripper and 8.3 for the hand; Seedance loses 8.8 and 4.4. Video Quality changes by only a few hundredths in every setting. The reference image therefore affects adherence to the robot appearance and source interaction in a model-dependent way, without materially changing presentation quality.

\subsection{Transfer Diagnostics}
\label{sec:diagnostic_failure_analysis}

\subsubsection{VBench vs. H2RCore.}

Figure~\ref{fig:m5_vs_h2rcore} compares the VBench-based Video Quality score with H2RCore across 22 model--embodiment pairs. Video Quality lies between 0.73 and 0.81, while H2RCore spans 30.0 to 84.6. Their rank association is weak (Spearman $\rho=0.14$). HunyuanVideo 1.5-I2V leads Video Quality score but sits near the bottom of H2RCore, whereas Mitty-EPIC14B has the lowest Video Quality score and substantially stronger overall scores. The comparison shows that Video Quality alone does not reliably recover the benchmark ranking across models.

Supplementary Figure~\ref{fig:diagnostic_failures} provides a more detailed breakdown of the discrepancy. Human-led manipulation and contact-region mismatch are prevalent among image/frame-conditioned outputs, while end-effector mismatch remains common for the Parallel-Jaw Gripper. The diagnostic categories distinguish failures caused by actor leakage, unsupported contact, and incorrect morphology.

\subsubsection{Qualitative Examples.}

Figure~\ref{fig:qualitative_metrics} compares three models on the same scoop-and-dump demonstration. Seedance 2.0 maintains visible robot contact with the scoop and transfers the ice into the cup. HunyuanVideo 1.5-I2V leaves the human as the active manipulator, preserving the action without transferring it to the robot. Veo 3.1 produces a plausible interaction with the wrong end effector. The four metrics capture complementary evidence: M1 checks the outcome, M2 the required events, M3 visible robot--scoop contact, and M4 the requested embodiment. A correct final state alone does not establish successful transfer.

\section{Conclusion}

We introduced H2R-Bench, a benchmark for evaluating whether video world models can transfer egocentric human manipulation demonstrations into videos of specified robot embodiments. Across six manipulation families and two target embodiments, H2R-Bench evaluates current video world models through five complementary dimensions, including task goals, action dynamics, functional contact, embodiment consistency, and video quality. Our evaluation of eleven representative video generation models reveals a substantial gap between visual generation quality and embodied transfer capability, suggesting that current video world models still struggle to capture the requirements of cross-embodiment manipulation transfer. We hope H2R-Bench will facilitate the development of video world models that bridge the human-to-robot embodiment gap, enabling the transformation of abundant human manipulation observations into reliable robot-centric training resources.

\IfFileExists{aaai2027.bib}{%
  \bibliography{aaai2027}%
}{%
  \bibliography{papers/aaai2027}%
}

\clearpage
\appendix
\setcounter{secnumdepth}{2}
\renewcommand{\thetable}{S\arabic{table}}
\renewcommand{\thefigure}{S\arabic{figure}}
\renewcommand{\theequation}{S\arabic{equation}}
\setcounter{table}{0}
\setcounter{figure}{0}
\setcounter{equation}{0}

\section{Dataset Construction and Annotation Quality Control}
\label{sec:dataset_annotation_qc}

\subsection{Source Selection and Task Stratification}
H2R-Bench is built from egocentric source clips selected from the EgoDex test split~\cite{hoque2025egodex} and balanced over six physical-state manipulation families. We retain clips only when the manipulated entities, a task-relevant state transition, and sufficient visual evidence for the interaction can be identified. Target embodiments are paired after source selection: the parallel-jaw gripper and dexterous-hand conditions therefore use exactly the same human source evidence.

The taxonomy is designed for the benchmark rather than inherited from EgoDex activity labels. It groups recurring embodied-manipulation problems by the physical state change that defines task completion, following the emphasis on diverse object interactions in robot-learning and egocentric-manipulation datasets~\cite{brohan2022rt,o2024open,hoque2025egodex}. The evaluation dimensions follow the evidence needed to determine whether that state change is reproduced by the requested robot. Table~\ref{tab:design_rationale} summarizes both design choices.

\begin{table*}[t]
\centering
\footnotesize
\renewcommand{\arraystretch}{1.08}
\setlength{\tabcolsep}{3pt}
\begin{tabular*}{\textwidth}{@{\extracolsep{\fill}}p{0.15\textwidth}p{0.22\textwidth}p{0.31\textwidth}p{0.22\textwidth}@{}}
\toprule
Element & Embodied-manipulation concern & H2R-Bench operationalization & Design origin \\
\midrule
\multicolumn{4}{@{}l}{\textit{Task families: task-defining physical state changes}} \\
F1: Rigid rearrangement & Object pose, support, or containment & Place or transport a rigid object into the demonstrated spatial relation. & Rigid transport and placement tasks. \\
F2: Mechanism actuation & State of an articulated mechanism & Open, close, toggle, press, or rotate a task-relevant mechanism. & Interaction with articulated objects. \\
F3: Insertion and assembly & Connection, fit, or attachment relation & Establish or remove a constrained connection between entities. & Precision alignment and constrained contact. \\
F4: Deformable configuration & Non-rigid shape or configuration & Produce the demonstrated fold, bend, compression, or shape change. & Deformable-object manipulation. \\
F5: Bulk-material transfer & Distribution or containment of material & Pour, scoop, transfer, or mix material between regions or containers. & Many-particle and material-flow manipulation. \\
F6: Surface/material transformation & Local surface condition or material integrity & Produce a visible local change through wiping, spreading, cutting, peeling, or related interaction. & Tool-mediated local transformation. \\
\midrule
\multicolumn{4}{@{}l}{\textit{Evaluation dimensions: evidence required for valid H2R generation}} \\
M1: Goal-State Completion & Was the demonstrated task state reached? & Score weighted predicates over the visible final state. & Task-success and outcome evaluation in robotic video benchmarks. \\
M2: Action-Event Completion & Were the required manipulation events shown? & Score completion of source-derived action events. & Action-completeness evaluation in robotic video benchmarks. \\
M3: Functional Contact Transfer & Did robot contact support the source-consistent object response? & Evaluate contact region, establishment, mode, temporal object response, and embodiment-compatible strategy. & Contact-mediated manipulation and H2R interaction grounding. \\
M4: Embodiment Correctness & Did the requested robot perform the manipulation? & Evaluate robot presence, human absence, embodiment category, end-effector subtype, and temporal structure. & Robot-structure and embodiment compliance. \\
M5: Video Quality & Is the generated video visually and temporally well formed? & Measure imaging quality, aesthetic quality, temporal stability, and motion smoothness. & Task-agnostic video-generation quality. \\
\bottomrule
\end{tabular*}
\caption{Design rationale for the H2R-Bench task taxonomy and evaluation dimensions. The six families are organized by task-defining physical state changes rather than semantic activity labels; M1--M5 cover task realization, source-relative interaction, target embodiment, and presentation quality.}
\label{tab:design_rationale}
\end{table*}

\subsection{Clip Grounding and Structured Annotation}
For each benchmark clip, Qwen3.7-Plus generates a structured annotation from the 5-second source video and 32 uniformly sampled frames, including the first and last frame. This annotation preserves functional information such as contact, support, release, and expected object response while avoiding a prescribed robot trajectory. Every annotation is manually checked against the source video. Corrections are incorporated before the annotations are used for evaluation.

\subsection{Annotation Acceptance Criteria}
This stage determines whether a source clip has sufficient evidence to support the benchmark evaluation. Each annotation is stored in a fixed JSON schema containing the task family, goal checks, required action events, interaction requirements, and embodiment strategies. We verify that these fields are complete and internally consistent before accepting a case.

Each case receives an explicit validity decision. We exclude clips when occlusion, ambiguous object states, or unclear interactions make the task goal or required contact impossible to assess reliably. For accepted cases, remaining minor ambiguities are recorded as notes for later analysis but do not change the evaluation protocol or official benchmark set.

\begin{table*}[t]
\centering
\small
\setlength{\tabcolsep}{4pt}
\begin{tabular}{p{0.20\linewidth}p{0.25\linewidth}p{0.47\linewidth}}
\toprule
Stage & Output & Quality control \\
\midrule
Source curation & Curated clips across six families & Identifiable manipulated entities, visible task-relevant state change, and sufficient interaction evidence. \\
Clip grounding & 32 ordered source frames per clip & First and last frames are retained; annotations identify initial state, contact, core transition, and final state when visible. \\
Structured annotation & Goals, events, milestones, interaction, and embodiment strategies & Strict JSON schema with canonical event identifiers and a fixed six-family taxonomy. \\
Completion and validation & Schema-complete case records & Only missing fields are completed; existing non-empty fields are preserved; schema and type mismatches are rejected. \\
Manual verification & All benchmark annotations & Every record is checked against its source video and corrected before evaluation. \\
Uncertainty accounting & Validity and ambiguity fields & Every record has a validity decision; notes document residual uncertainty without changing the official evaluation set. \\
\bottomrule
\end{tabular}
\caption{Dataset construction and annotation quality controls. Source-side annotations define the benchmark specification; detailed frame-level evidence and uncertainty fields are not exposed to generation models.}
\label{tab:dataset_annotation_qc}
\end{table*}

\section{Additional Experimental Details}

\subsection{Diagnostic Failure Rates}
\label{sec:diagnostic_failure_rates_appendix}

\begin{figure*}[t]
\centering
\includegraphics[width=0.98\textwidth]{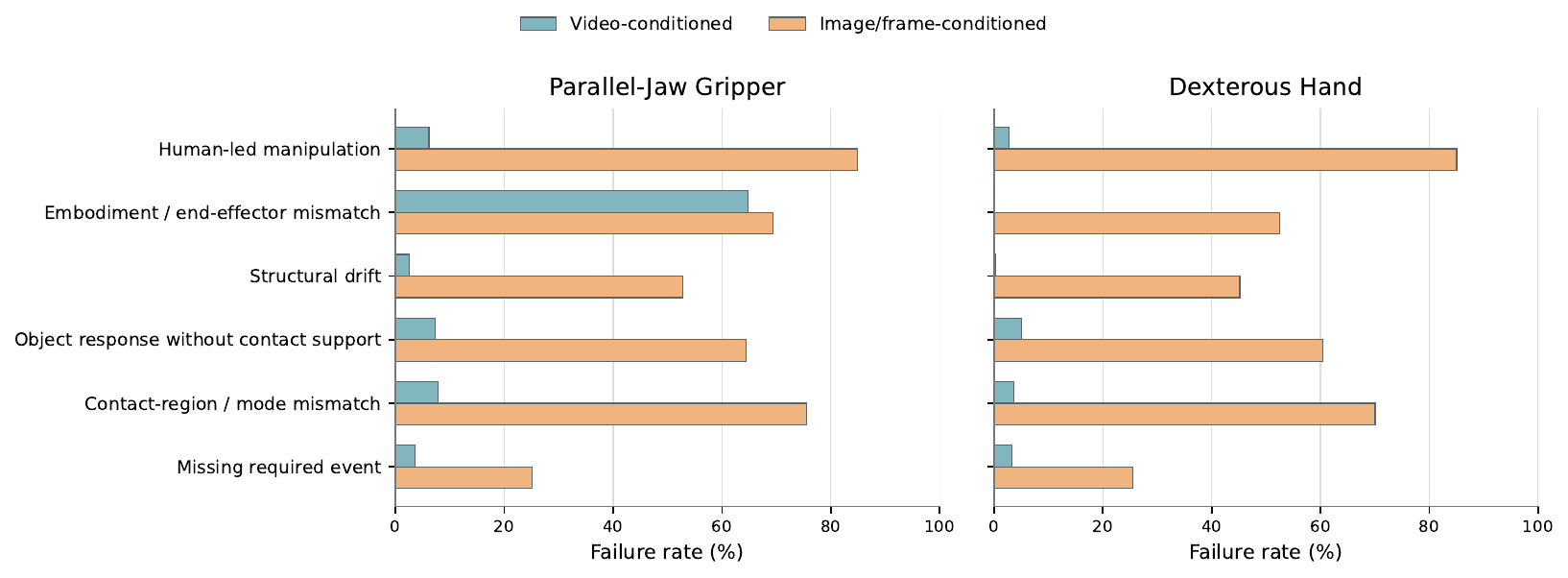}
\caption{Failure-type rates derived from structured M2--M4 diagnostics on the 120-source main evaluation set. Each bar reports the percentage of evaluated videos in the corresponding conditioning group whose judge-averaged diagnostic component is below $0.5$. Categories are non-exclusive. Video-conditioned models show lower observed rates of human-led manipulation and contact failures, whereas Parallel-Jaw Gripper transfer remains sensitive to end-effector mismatch and both embodiments retain action-event failures.}
\label{fig:diagnostic_failures}
\end{figure*}

On the 120-source main evaluation set, Figure~\ref{fig:diagnostic_failures} aggregates diagnostic failures identified from the structured M2--M4 outputs. A failure is recorded when the corresponding judge-averaged diagnostic component is below $0.5$; categories are non-exclusive because one generation can violate embodiment, contact, and action requirements simultaneously. In this native-interface collection, image/frame-conditioned systems frequently retain human-led manipulation and exhibit weak visible contact support or an incorrect contact region or mode. The three video-conditioned systems have lower observed failure rates, but this descriptive comparison does not identify an interface effect because model and interface vary together. Parallel-Jaw Gripper transfer still shows frequent end-effector mismatch, while required action events are sometimes omitted for both target embodiments. These patterns demonstrate the value of reporting the transfer components alongside H2RCore: the benchmark identifies whether a failure arises from action completion, functional contact, or target embodiment rather than reducing all errors to a single quality score.

M3 additionally treats source-scene or task-entity substitution as a hard failure: generic-looking contact in a different scene is not contact transfer from the source demonstration. The source-grounding decision is made independently for each generated video before the contact dimensions are aggregated.

\subsection{Matched Source-Conditioning Ablation}

To test whether sparse image evidence can substitute for a source video, we compare Seedance 2.0's native video-conditioned run with a nine-image setting that provides nine chronologically ordered, uniformly sampled source frames. This ablation uses 24 matched sources from the main evaluation set, with four sources from each task family. Both target embodiments are evaluated for every source, giving 48 transfer cases. Replacing the complete source video with nine images preserves much of the final-state and broad action evidence, but substantially degrades functional contact and embodiment realization. H2RCore falls by 27.6 points for the Parallel-Jaw Gripper and 41.4 points for the Dexterous Hand, while Video Quality changes by less than 0.02 in both settings (Table~\ref{tab:source_conditioning_ablation}). Qualitative inspection and M4 diagnoses show that the nine-image interface frequently reproduces the human manipulation rather than reliably replacing the human actor with the requested robot.

\begin{table}[t]
\centering
\footnotesize
\setlength{\tabcolsep}{1.7pt}
\begin{tabular}{@{}llcccccc@{}}
\toprule
Target & Input & M1 & M2 & M3 & M4 & M5 & Core \\
\midrule
\multirow{2}{*}{\shortstack{Parallel-Jaw\\Gripper}} & Video & 0.737 & 0.850 & 0.709 & 0.598 & 0.793 & 70.9 \\
 & 9 images & 0.744 & 0.848 & 0.284 & 0.104 & 0.783 & 43.4 \\
\midrule
\multirow{2}{*}{\shortstack{Dexterous\\Hand}} & Video & 0.768 & 0.857 & 0.861 & 0.898 & 0.799 & 85.1 \\
 & 9 images & 0.753 & 0.856 & 0.313 & 0.079 & 0.782 & 43.7 \\
\bottomrule
\end{tabular}
\caption{Matched source-conditioning ablation for Seedance 2.0 on 24 sources. Both target embodiments are evaluated for every source. ``Video'' uses the full source clip; ``9 ordered images'' uses uniformly sampled chronological frames.}
\label{tab:source_conditioning_ablation}
\end{table}

\subsection{Performance Across Task Families}
\label{sec:family_breakdown_appendix}

\begin{figure*}[t]
\centering
\includegraphics[width=0.98\textwidth]{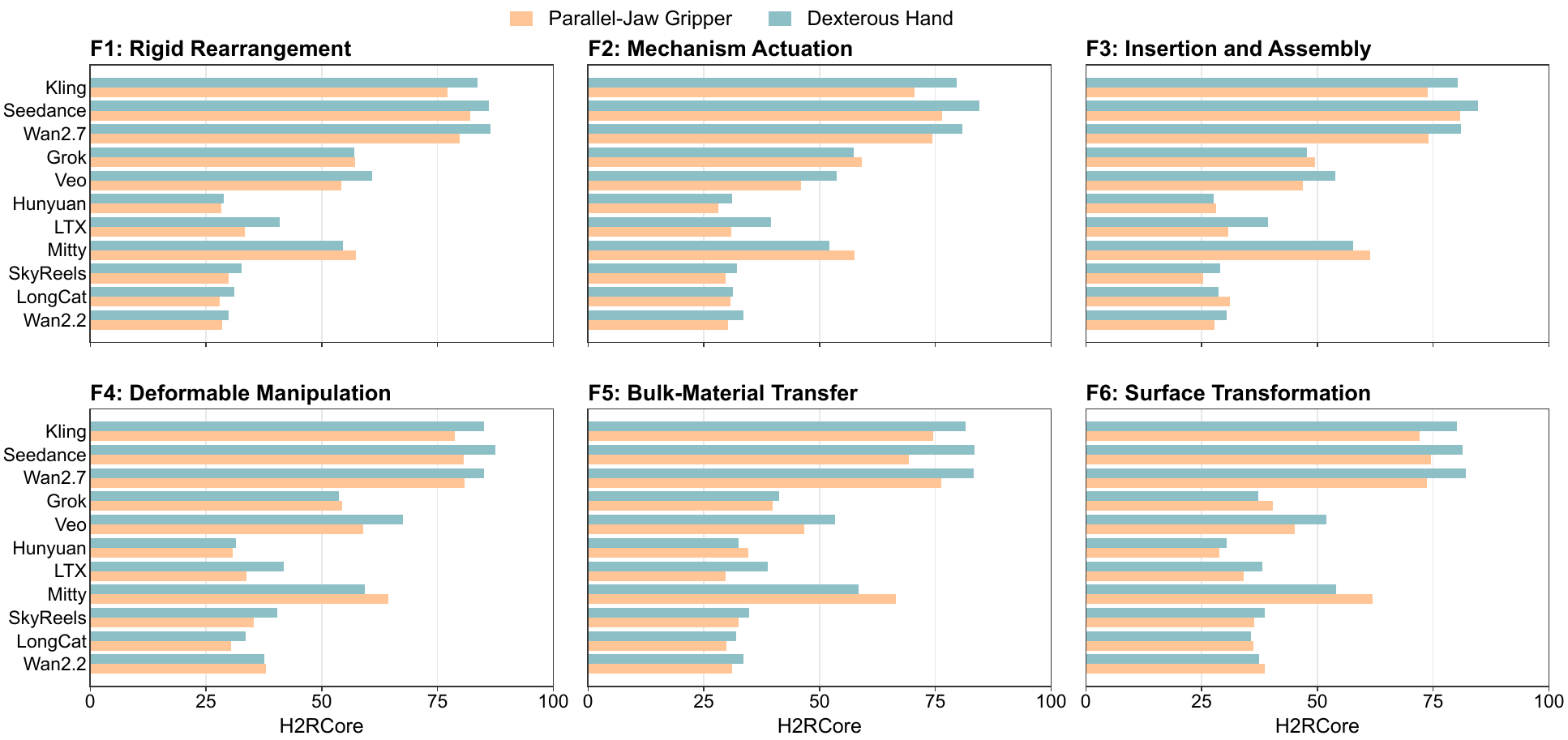}
\caption{Task-family H2RCore profiles across the six manipulation families. Each panel compares Parallel-Jaw Gripper and Dexterous Hand transfer for the same 11 models. Higher is better, and all panels share a 0--100 scale, exposing both task-dependent difficulty and embodiment-dependent model preferences.}
\label{fig:taskwise_h2rcore}
\end{figure*}

On the 120-source main evaluation set, aggregate scores conceal substantial task dependence. Figure~\ref{fig:taskwise_h2rcore} separates H2RCore by manipulation family and target embodiment. Averaged over models, deformable-object configuration (F4) remains the strongest family for both targets, while the other interaction families exhibit lower aggregate transfer scores. This ordering suggests that difficulty is governed less by visible deformation itself than by whether success requires precise, localized functional contact and a partially hidden state transition. The panels also expose model-specific preferences: Seedance and Wan2.7 are comparatively stable across families, whereas Grok Imagine Video and Veo 3.1 show substantially larger task-dependent variation. A model can therefore appear competitive on visible final states in selected tasks while remaining unreliable for a particular interaction class.

Table~\ref{tab:family_breakdown} provides the per-task-family breakdown on the main evaluation set for the six manipulation families. Each family stresses a different aspect of human-to-robot transfer: rigid transport, mechanism actuation, insertion and assembly, deformable configuration change, bulk-material transfer, and surface transformation. Parallel-Jaw Gripper and Dexterous Hand targets are reported in separate column groups, and row colors identify the task families.

\subsection{Attribute-Based Analysis}
\label{sec:attribute_analysis}

\begin{figure*}[t]
\centering
\includegraphics[width=0.78\textwidth]{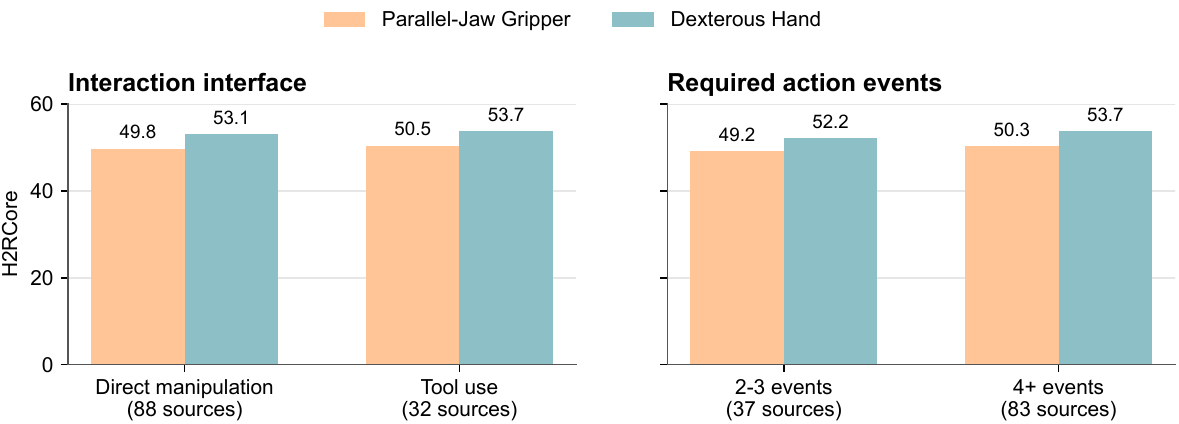}
\caption{H2RCore comparison across annotation-derived task attributes on the 120-source main evaluation set. The left panel separates direct manipulation from tool use; the right groups tasks by the number of required action events. Bars show mean H2RCore over the evaluated model outputs for each target embodiment, values are printed above the bars, and parenthetical counts give the number of source scenes in each attribute group. Both panels use the same vertical scale.}
\label{fig:attribute_breakdown}
\end{figure*}

Using the same 120-source main evaluation set, task-family averages describe the dominant physical state change, but they do not expose whether difficulty is associated with the interaction interface or the number of required actions. Figure~\ref{fig:attribute_breakdown} therefore groups the evaluations using annotation fields available across the benchmark: direct manipulation versus tool use, and tasks with two--three versus four or more required events. The splits reveal systematic variation that family-level means can conceal. They are descriptive diagnostics rather than causal comparisons, because task attributes naturally co-occur.

\subsection{Statistical Reporting}
\label{sec:statistical_reporting}

Table~\ref{tab:h2rcore_ci} reports percentile 95\% confidence intervals (CIs) for H2RCore on the 120-source main evaluation set. The intervals are obtained by stratified bootstrap resampling of source tasks within each task family.

\begin{table}[!t]
\centering
\footnotesize
\renewcommand{\arraystretch}{0.98}
\setlength{\tabcolsep}{3pt}
\begin{tabular}{lcc}
\toprule
Model & \shortstack{Parallel-Jaw Gripper\\Core [95\% CI]} & \shortstack{Dexterous Hand\\Core [95\% CI]} \\
\midrule
Kling-V3 & 74.5 [73.6, 75.3] & 81.7 [80.8, 82.6] \\
Seedance 2.0 & 77.3 [76.5, 78.1] & 84.6 [84.0, 85.3] \\
Wan2.7 & 76.5 [75.7, 77.4] & 83.1 [82.3, 83.9] \\
Grok Imagine & 50.1 [49.0, 51.1] & 49.2 [48.3, 50.1] \\
Veo 3.1 & 49.6 [48.9, 50.4] & 57.0 [56.1, 57.8] \\
Hunyuan-1.5 & 30.0 [28.8, 31.2] & 30.7 [29.5, 31.8] \\
LTX-2.3 & 32.1 [31.0, 33.2] & 39.8 [38.6, 40.9] \\
Mitty-EPIC & 61.5 [60.5, 62.5] & 56.1 [55.2, 57.0] \\
SkyReels-V3 & 31.5 [30.5, 32.6] & 34.6 [33.6, 35.6] \\
LongCat & 31.0 [29.9, 32.1] & 32.0 [30.9, 33.2] \\
Wan2.2 & 32.4 [31.4, 33.3] & 33.7 [32.7, 34.7] \\
\bottomrule
\end{tabular}
\caption{Main-evaluation H2RCore point estimates and percentile 95\% confidence intervals. Intervals summarize variation across source tasks under the five-metric aggregation rule.}
\label{tab:h2rcore_ci}
\end{table}

\begin{table}[!t]
\centering
\footnotesize
\renewcommand{\arraystretch}{1.02}
\setlength{\tabcolsep}{1.5pt}
\begin{tabular}{@{}lcc@{}}
\toprule
Comparison & \shortstack{Gripper\\$\Delta$ Core [95\% CI]} & \shortstack{Hand\\$\Delta$ Core [95\% CI]} \\
\midrule
Seedance $-$ Wan2.7 & $+0.79$ [$+0.08$, $+1.48$] & $+1.52$ [$+0.92$, $+2.16$] \\
Seedance $-$ Kling & $+2.82$ [$+2.04$, $+3.61$] & $+2.89$ [$+2.15$, $+3.73$] \\
Wan2.7 $-$ Kling & $+2.04$ [$+1.40$, $+2.66$] & $+1.37$ [$+0.66$, $+2.16$] \\
\bottomrule
\end{tabular}
\caption{Paired source-task H2RCore differences for the leading video-conditioned models. Positive values favor the left model; percentile 95\% CIs pair the same source tasks.}
\label{tab:leading_pairwise_ci}
\end{table}

\paragraph{Interpreting paired comparisons.}
All displayed paired intervals exclude zero under the stated bootstrap procedure. The smallest separation is Seedance--Wan2.7 for the Parallel-Jaw Gripper target.

\paragraph{Weight sensitivity.}
We recompute embodiment-specific rankings using equal weights $(0.20,0.20,0.20,0.20,0.20)$, a larger quality weight $(0.15,0.15,0.25,0.25,0.20)$, and a transfer-only variant $(0.15,0.15,0.35,0.35,0)$. Kendall rank correlation with the default ranking is between $0.927$ and $1.000$ across these settings, indicating that the aggregate ordering is stable under these alternatives.

\subsection{Detailed Evaluation Protocol}
\label{sec:detailed_evaluation}

All MLLM-based metrics use 25 uniformly sampled frames from each input video and expose only the annotation fields required for the metric. Judges return structured JSON containing per-item scores, evidence-frame indices, and rationales; scripts normalize scores and perform every aggregation. Let $c$ denote a generated video, $m\in\mathcal{J}$ one of the three judges, and $[z]_4=z/4$ the normalization of an integer score $z\in\{0,1,2,3,4\}$. A score of $0$ denotes absent or contradicted evidence, $1$ weak evidence, $2$ partial completion, $3$ mostly correct evidence with a minor defect or ambiguity, and $4$ clear, complete, and stable evidence.

\paragraph{Judge configuration.}
M1--M4 use Gemini 3.5 Flash (\texttt{gemini-3.5-flash}), Qwen3.7-Plus (\texttt{qwen3.7-plus}), and GPT-5.4 (\texttt{gpt-5.4}) at temperature 0. The three judges contribute equally to each MLLM-based metric.

\paragraph{M1: Goal-State Completion.}
Each case defines a set of weighted final-state predicates $\mathcal{G}_c$. A predicate may specify a spatial relation, mechanism state, attachment relation, material distribution, deformation, or local surface change. Judge $m$ receives 25 uniformly sampled generated frames and assigns $r^m_g\in\{0,\ldots,4\}$ to every $g\in\mathcal{G}_c$ based on the state visible at the end of the sequence. Thus, M1 evaluates the final state without discarding the preceding evidence that shows whether it was reached and preserved. With annotation weight $w_g$, the per-judge score is
\begin{equation}
S^m_{\mathrm{goal}}=
\frac{\sum_{g\in\mathcal{G}_c}w_g[r^m_g]_4}
     {\sum_{g\in\mathcal{G}_c}w_g}.
\end{equation}
We additionally report weighted predicate coverage at strict thresholds,
\begin{equation}
C^m_{\mathrm{goal}}(\tau)=
\frac{\sum_{g\in\mathcal{G}_c}w_g\mathbf{1}[r^m_g\geq\tau]}
     {\sum_{g\in\mathcal{G}_c}w_g},
\qquad \tau\in\{3,4\},
\end{equation}
which distinguishes partially reached outcomes from clearly completed ones.

\paragraph{M2: Action-Event Completion.}
Each annotation contains weighted required events $\mathcal{E}_c$. Judges inspect the 25 uniformly sampled generated frames and assign every event $e$ a completion score $a^m_e\in\{0,\ldots,4\}$. Event completion is the weighted mean
\begin{equation}
S^m_{\mathrm{action}}=
\frac{\sum_{e\in\mathcal{E}_c}w_e[a^m_e]_4}
     {\sum_{e\in\mathcal{E}_c}w_e}.
\end{equation}
We also report weighted event coverage at scores $\geq3$ and $=4$. M2 summarizes the visible completion of the required action events across the generated sequence.

\paragraph{M3: Functional Contact Transfer.}
M3 evaluates whether the robot realizes the source interaction functionally rather than merely reproducing a similar object trajectory. A contact specification is derived from the source annotation and 25 uniformly sampled source frames. It identifies manipulated entities, functional contact regions, manipulation modes, expected object responses, support or bimanual roles when relevant, and target-embodiment constraints. The judge compares this specification and the 25 source frames with 25 generated frames along five dimensions: (i) \emph{contact-region transfer}, whether the robot acts on the same or a functionally equivalent region; (ii) \emph{contact establishment}, whether visible contact is sustained enough to control the object; (iii) \emph{manipulation-mode transfer}, whether the robot uses a compatible functional mode such as grasping, pinching, pushing, pulling, stabilizing, or scooping; (iv) \emph{temporally supported object response}, whether the expected state change visibly follows robot contact or control; and (v) \emph{embodiment-compatible contact strategy}, whether the visible strategy is compatible with the requested gripper or dexterous hand. The fourth item is visual-temporal evidence, not a claim that contact physically caused the response; M4 separately evaluates the robot's visible structure and identity.

Before these dimensions are scored, the judge verifies source grounding: the generated video must preserve the source scene and camera context together with task-relevant objects, tools, and containers, allowing only task-required state changes and embodiment replacement. A substantial scene or task-entity substitution is a hard failure and receives zero M3 credit; this rule prevents generic contact in a different scene from being counted as source-contact transfer.

Applicability is specified by the case annotation and is shared by all judges. Let $\mathcal{C}_c$ denote this fixed subset of dimensions and let $h^m_q\in\{0,\ldots,4\}$ be judge $m$'s score for dimension $q$. A dimension marked as non-applicable is excluded for every judge rather than assigned a neutral score. The per-judge M3 score is
\begin{equation}
S^m_{\mathrm{contact}}=
\begin{cases}
0, & \text{if source grounding fails},\\
\displaystyle\frac{1}{|\mathcal{C}_c|}
\sum_{q\in\mathcal{C}_c}[h^m_q]_4, & \text{otherwise}.
\end{cases}
\end{equation}
Evidence indices are validated independently within the source and generated 25-frame sequences; consequently, source and generated frame identifiers cannot be conflated during judging.

\paragraph{M4: Embodiment Correctness.}
M4 uses 25 uniformly sampled generated frames and the requested embodiment specification only. Judges score five dimensions: robot-actor presence ($p$), absence of human hands or arms ($u$), broad embodiment-category match ($b$), end-effector correctness ($e$), and structural consistency over time ($s$). The presence term assesses visible robotic embodiment irrespective of subtype; category and end-effector are deliberately separate, so a visible dexterous hand can receive presence credit while receiving low morphology credit when the parallel-jaw gripper is requested. Let $p^m,u^m,b^m,e^m,s^m\in\{0,\ldots,4\}$ denote the five scores from judge $m$. A judge-marked hard failure, triggered when no robotic actor is visible or when human hands perform the principal manipulation, receives zero. Otherwise, the per-judge score is
\begin{equation}
\begin{aligned}
S^m_{\mathrm{emb}}={}&
0.20[p^m]_4+0.20[u^m]_4+0.20[b^m]_4\\
&+0.25[e^m]_4+0.15[s^m]_4.
\end{aligned}
\end{equation}
The larger end-effector weight reflects that the distinction between the parallel-jaw gripper and dexterous hand is central to the benchmark, while the hard-failure rule prevents a stable but incorrect human actor from receiving credit through the remaining dimensions.

\paragraph{M5: Video Quality.}
M5 is intentionally task-agnostic. Let $I_1,\ldots,I_T$ be the decoded frames of generated video $c$, and let $\operatorname{MAE}$ average absolute RGB differences over all pixels and channels. Framewise components traverse the decoded sequence, while temporal components use adjacent pairs or interpolation triplets. For MUSIQ, frames retain their aspect ratio and are downsampled only when the longer side exceeds 512 pixels; the aesthetic predictor uses the standard 224-pixel CLIP center crop. Imaging quality is the mean framewise output of the MUSIQ model trained on SPAQ~\cite{ke2021musiq}, normalized by 100:
\begin{equation}
v_{c,\mathrm{IQ}}=\frac{1}{100T}\sum_{t=1}^{T}
\operatorname{MUSIQ}(I_t).
\end{equation}
Aesthetic quality is the mean output of the LAION linear aesthetic predictor applied to normalized CLIP ViT-L/14 image features~\cite{radford2021clip,laion2022aesthetic}, normalized by 10:
\begin{equation}
v_{c,\mathrm{AQ}}=\frac{1}{10T}\sum_{t=1}^{T}
\operatorname{LAION}\!\left(
\frac{\operatorname{CLIP}(I_t)}
{\lVert\operatorname{CLIP}(I_t)\rVert_2}\right).
\end{equation}
Temporal stability is one minus the mean normalized difference between consecutive frames:
\begin{equation}
v_{c,\mathrm{TF}}=
1-\frac{1}{255(T-1)}
\sum_{t=1}^{T-1}\operatorname{MAE}(I_t,I_{t+1}).
\end{equation}
For motion smoothness, AMT-S~\cite{li2023amt} predicts each odd frame $\widehat I_{2k+1}$ from its two neighboring even frames at interpolation time $0.5$. With $K$ valid triplets,
\begin{equation}
v_{c,\mathrm{MS}}=
1-\frac{1}{255K}\sum_{k=0}^{K-1}
\operatorname{MAE}(I_{2k+1},\widehat I_{2k+1}).
\end{equation}
Each component is clipped to $[0,1]$ before the four values are averaged:
\begin{equation}
S_{\mathrm{video},c}=
\frac{1}{4}\sum_{q\in\{\mathrm{IQ,AQ,TF,MS}\}}v_{c,q}.
\end{equation}
M5 therefore measures the presentation quality of a video without rewarding an incorrect manipulation or embodiment.

\paragraph{Judge, dataset, and overall aggregation.}
For $X\in\{\mathrm{goal},\mathrm{action},\mathrm{contact},\mathrm{emb}\}$, we average the three judge scores for each video and then average across the evaluated set $\mathcal{D}$:
\begin{equation}
S_X=\frac{1}{|\mathcal{D}|}\sum_{c\in\mathcal{D}}
\left(\frac{1}{|\mathcal{J}|}\sum_{m\in\mathcal{J}}S^m_{X,c}\right),
\qquad |\mathcal{J}|=3.
\end{equation}
All tables retain the component scores because they diagnose distinct failure modes. H2RCore aggregates the five dataset-level component scores:
\begin{equation}
\begin{aligned}
\mathrm{H2RCore}=100\bigl(&
0.15S_{\mathrm{goal}}+0.15S_{\mathrm{action}}\\
&+0.30S_{\mathrm{contact}}+0.30S_{\mathrm{emb}}\\
&+0.10S_{\mathrm{video}}\bigr).
\end{aligned}
\end{equation}
This aggregation prioritizes contact transfer and embodiment correctness, while assigning a smaller contribution to Video Quality. For human-agreement analyses, where raters score M1--M4 but not M5, we use the transfer subscore $100(0.20S_{\mathrm{goal}}+0.20S_{\mathrm{action}}+0.30S_{\mathrm{contact}}+0.30S_{\mathrm{emb}})$.

\subsection{Human--Automatic Agreement}
\label{sec:human_automatic_agreement}

Three human raters score M1--M4 on the same random sample of 660 videos described in the main paper. We compare the final human scores with the averaged MLLM scores for records with complete paired evaluations. Pearson correlation is computed across videos for each transfer metric. The corresponding coefficients are 0.791 for M1, 0.818 for M2, 0.880 for M3, and 0.877 for M4 (Table~\ref{tab:human_automatic_pearson}). Combining M1--M4 with the transfer-subscore weights defined above gives $r=0.930$. Within-scene model-ranking agreement is reported separately in Figure~\ref{fig:human_mllm_spearman}.

\begin{table}[!ht]
\centering
\footnotesize
\setlength{\tabcolsep}{2.5pt}
\begin{tabular*}{\columnwidth}{@{\extracolsep{\fill}}lccccc@{}}
\toprule
Score & M1 & M2 & M3 & M4 & \shortstack{Transfer\\Subscore} \\
\midrule
Pearson $r$ & 0.791 & 0.818 & 0.880 & 0.877 & 0.930 \\
\bottomrule
\end{tabular*}
\caption{Human--automatic agreement for paired human and MLLM evaluations. M1--M4 are compared per video; the transfer subscore combines these four metrics without M5.}
\label{tab:human_automatic_pearson}
\end{table}

\subsection{Inter-Judge Agreement}
\label{sec:inter_judge_agreement}

We assess agreement among Gemini, Qwen, and GPT on the 120-source main evaluation set for M1--M4. For each metric and judge pair $(p,q)$, we compute the Pearson correlation $r_{pq}$ between their per-video scores. Scores are computed from the structured per-item outputs using the arithmetic definitions in Section~\ref{sec:detailed_evaluation}; M5 is excluded because it does not use MLLM judges.

\begin{table}[!ht]
\centering
\footnotesize
\setlength{\tabcolsep}{1.2pt}
\begin{tabular*}{\columnwidth}{@{\extracolsep{\fill}}lcccc@{}}
\toprule
Judge Pair & M1 & M2 & M3 & M4 \\
\midrule
Gemini / Qwen & 0.695 & 0.691 & 0.741 & 0.874 \\
Gemini / GPT & 0.582 & 0.636 & 0.726 & 0.891 \\
Qwen / GPT & 0.656 & 0.690 & 0.809 & 0.892 \\
\bottomrule
\end{tabular*}
\caption{Pairwise inter-judge agreement on the 120-source main evaluation set. Each cell reports Pearson correlation $r$ between per-video scores.}
\label{tab:inter_judge_agreement}
\end{table}

\begin{table}[t]
\centering
\footnotesize
\setlength{\tabcolsep}{2pt}
\begin{tabular*}{\columnwidth}{@{\extracolsep{\fill}}lll@{}}
\toprule
Model & Interface & Source input \\
\midrule
Seedance 2.0 & Video & Full clip \\
Wan2.7 & Video & Full clip \\
Kling-V3 & Video & Full clip \\
Veo 3.1 & Frame & First + last \\
Grok Imagine Video & Frame & Up to 7 frames \\
HunyuanVideo 1.5-I2V & Frame & Ordered frame(s) \\
LTX-2.3 & Frame & Ordered frame(s) \\
Mitty-EPIC14B & Frame & Ordered frame(s) \\
SkyReels-V3-R2V & Frame & Ordered frame(s) \\
LongCat & Frame & Ordered frame(s) \\
Wan2.2 & Frame & First frame \\
\bottomrule
\end{tabular*}
\caption{Source-conditioning interfaces in the main experiment. Frame inputs are sampled in temporal order from the source demonstration. No model receives a target-robot reference image.}
\label{tab:model_inputs}
\end{table}
\section{Model Descriptions and Implementation Setups}
\label{sec:model_implementation_appendix}

We evaluate 11 representative video generators through their publicly available hosted interfaces or local implementations. This section records the generation configuration used by H2R-Bench rather than attempting to equate heterogeneous provider-side sampling controls. The main comparison uses the strongest documented source-conditioning interface available for each model. All systems receive an aligned case-specific English prompt that specifies the task outcome and either a parallel-jaw gripper or a dexterous hand; its presentation is adapted only to the input limits of the corresponding interface. The prompt does not expose evaluation-only annotations. Unless stated otherwise, no target-robot appearance reference is supplied in the main experiment; target-reference results are reported separately in Table~\ref{tab:robot_reference_ablation}.

\subsection{Computing Infrastructure}

Locally hosted generation, video preprocessing, M5 computation, and result aggregation were performed on a workstation running Ubuntu 24.04.2 LTS with two Intel Xeon Gold 6144 processors (16 physical cores and 32 hardware threads), 251 GiB of system memory, and four NVIDIA GeForce RTX 4090 GPUs with 24 GiB of reported memory per GPU. The system used NVIDIA driver 580.173.02, CUDA 13.0, and cuDNN 9.19.0. The software environment used Python 3.10.16, PyTorch 2.11.0, torchvision 0.26.0, Transformers 4.57.1, Diffusers 0.36.0, Accelerate 1.12.0, OpenCV 4.13.0, NumPy 2.2.6, SciPy 1.15.3, pandas 2.3.3, scikit-learn 1.5.0, Decord 0.6.0, and FFmpeg 6.1.1. Commercial video generators and the three MLLM judges were accessed through provider-hosted APIs; their server-side hardware is not disclosed to us.

\subsection{Commercially Hosted Models}

\textbf{Seedance 2.0.} We use the official Volcengine Ark interface in its video-conditioned mode. Each request provides the complete source clip through the supported public-video interface together with the case-specific target-embodiment prompt. The main setting uses no robot reference image.

\textbf{Wan2.7.} We use the provider's video-conditioned interface, supplying the source demonstration and the common target-embodiment prompt. This setting tests direct retargeting from temporally ordered human manipulation evidence without a target-robot image.

\textbf{Kling-V3.} We use Kling's video-conditioned generation interface with the full source demonstration and the common prompt. The requested robot morphology is specified in text only in the main experiment.

\textbf{Veo 3.1.} We use the supported image-conditioned interface, providing the first and last source frames in temporal order rather than the source video. The prompt identifies the task and requested embodiment; no target-robot reference image is included.

\textbf{Grok Imagine Video.} We use the image-conditioned interface with multiple source frames sampled in temporal order (up to seven uniformly spaced frames when supported). The model receives the common task and embodiment prompt and no robot appearance reference.

\subsection{Open-Source or Locally Hosted Models}
\textbf{HunyuanVideo 1.5-I2V.} We run the image-to-video implementation with source frame conditioning and the target-embodiment prompt. The source video itself is not passed to the model.

\textbf{LTX-2.3.} We use the available image/frame-conditioned generation mode, supplying source frame evidence and the common prompt. We retain the provider output for evaluation rather than rendering a target robot externally.

\textbf{Mitty-EPIC14B.} We use the locally hosted image/frame-conditioned implementation with source frame inputs and the same target-embodiment prompt used across models.

\textbf{SkyReels-V3-R2V.} We use the available image/frame-conditioned interface with temporally ordered source frames and the shared prompt.

\textbf{LongCat.} We use the locally hosted image/frame-conditioned configuration with source frame inputs and the shared prompt. No target-robot reference image is supplied.

\textbf{Wan2.2.} We use the local Wan2.2-TI2V-5B implementation. Its one-image input budget is filled by the first frame of the source demonstration; the full source clip is not supplied. Parallel-Jaw Gripper and Dexterous Hand generations use their corresponding case prompts and the same seed for a paired source clip. The generation profile produces 121 frames at 24 fps (approximately 5 seconds) and preserves the source aspect ratio within the model's 720P-area setting.

\subsection{Visual Inputs and Output Profiles}
\label{sec:generation_profiles}

For every frame-conditioned interface, source images are kept in chronological order. When an interface accepts one source image, we use the first source frame; when it accepts multiple source images, we use uniformly spaced frames including the temporal endpoints, up to the interface input budget. Veo 3.1 receives the first and last frames, and Grok Imagine Video receives up to seven uniformly spaced frames. The video-conditioned systems receive the complete source clip. Table~\ref{tab:generation_profiles} reports the decoded profiles of the outputs retained for evaluation. These native profiles are not normalized before generation; comparability is instead established by the fixed evaluation frame budgets described below.

\begin{table}[t]
\centering
\scriptsize
\setlength{\tabcolsep}{2pt}
\begin{tabular*}{\columnwidth}{@{\extracolsep{\fill}}lcc@{}}
\toprule
Model & Decoded Size & Frames @ fps \\
\midrule
Seedance 2.0 & $1280\times720$ & 121 @ 24 \\
Wan2.7 & $1280\times720$ & 150 @ 30 \\
Kling-V3 & $1280\times720$ & 121 @ 24 \\
Veo 3.1 & $1280\times720$ & 144 @ 24 \\
Grok Imagine Video & $1280\times720$ & 121 @ 24 \\
HunyuanVideo 1.5-I2V & $1280\times720$ & 121 @ 24 \\
LTX-2.3 & $1280\times720$ & 120 @ 24 \\
Mitty-EPIC14B & $1280\times720$ & 37 @ 8 \\
SkyReels-V3-R2V & $1280\times720$ & 121 @ 24 \\
LongCat & $832\times480$ & 80 @ 15 \\
Wan2.2 & 720P area & 121 @ 24 \\
\bottomrule
\end{tabular*}
\caption{Native decoded output profiles in the main experiment. ``Frames @ fps'' reports decoded frame count and frame rate. Wan2.2 preserves source aspect ratio within its 720P-area setting; audio is removed before evaluation.}
\label{tab:generation_profiles}
\end{table}

\subsection{Output Handling}
We retain completed provider or local outputs in their native generation format and do not post-process clips to insert, render, or remove an actor. For comparable evaluation, every input video is decoded and uniformly sampled into 25 frames for M1--M4; M3 uses one 25-frame sequence from the source and one from the generated video. M5 is computed on the generated video itself. Thus, the interface difference is explicit at generation time while the evaluation evidence budget is fixed across models. Tables~\ref{tab:model_inputs} and~\ref{tab:generation_profiles} summarize the source-conditioning interfaces and native output profiles.
\subsection{Prompt Examples}
\label{sec:prompt_examples}

The visual condition is supplied through each model's native interface, while the text specifies the shared task, scene, contact, and embodiment constraints. Figure~\ref{fig:prompt_examples} presents abridged generation prompts for the same dexterous-hand case, the embodiment clause, the structured annotation input, and an M3 judge example. Source-video annotations are evaluation-only and are not exposed to the video generators. Complete case-specific prompts, annotation templates, judge prompts, and JSON schemas are included in the benchmark release.

\begin{table*}[t]
\centering
\scriptsize
\renewcommand{\arraystretch}{0.88}
\setlength{\tabcolsep}{0.8pt}
\begin{tabular}{clrrrrrrrrrrrr}
\toprule
 & & \multicolumn{6}{c}{Parallel-Jaw Gripper} & \multicolumn{6}{c}{Dexterous Hand} \\
\cmidrule(lr){3-8}\cmidrule(lr){9-14}
Task & Model & \shortstack{Goal\\Completion} & \shortstack{Action\\Completion} & \shortstack{Contact\\Transfer} & \shortstack{Embod.\\Correct.} & \shortstack{Video\\Quality} & Core & \shortstack{Goal\\Completion} & \shortstack{Action\\Completion} & \shortstack{Contact\\Transfer} & \shortstack{Embod.\\Correct.} & \shortstack{Video\\Quality} & Core \\
\midrule
\rowcolor{h2rfamilyone}  & Kling-V3 & 0.736 & 0.837 & 0.782 & 0.735 & 0.803 & 77.1 & 0.738 & 0.837 & 0.847 & 0.886 & 0.807 & 83.7 \\
\rowcolor{h2rfamilyone}  & Seedance 2.0 & 0.762 & 0.857 & 0.836 & 0.824 & 0.798 & 82.1 & 0.775 & 0.853 & 0.880 & 0.908 & 0.802 & 86.1 \\
\rowcolor{h2rfamilyone}  & Wan2.7 & 0.744 & 0.843 & 0.824 & 0.776 & 0.793 & 79.7 & 0.777 & 0.845 & 0.890 & 0.913 & 0.799 & 86.4 \\
\rowcolor{h2rfamilyone}  & Grok Imagine & 0.608 & 0.704 & 0.555 & 0.430 & 0.806 & 57.3 & 0.656 & 0.742 & 0.561 & 0.368 & 0.811 & 57.0 \\
\rowcolor{h2rfamilyone}  & Veo 3.1 & 0.751 & 0.791 & 0.574 & 0.197 & 0.790 & 54.2 & 0.780 & 0.830 & 0.653 & 0.305 & 0.800 & 60.9 \\
\rowcolor{h2rfamilyone}  & Hunyuan-1.5 & 0.520 & 0.446 & 0.188 & 0.000 & 0.812 & 28.2 & 0.483 & 0.429 & 0.209 & 0.027 & 0.814 & 28.9 \\
\rowcolor{h2rfamilyone}  & LTX-2.3 & 0.424 & 0.550 & 0.344 & 0.018 & 0.784 & 33.3 & 0.455 & 0.558 & 0.422 & 0.170 & 0.791 & 40.9 \\
\rowcolor{h2rfamilyone}  & Mitty-EPIC & 0.522 & 0.591 & 0.540 & 0.575 & 0.733 & 57.5 & 0.528 & 0.640 & 0.591 & 0.402 & 0.733 & 54.6 \\
\rowcolor{h2rfamilyone}  & SkyReels-V3 & 0.408 & 0.527 & 0.267 & 0.000 & 0.792 & 29.9 & 0.376 & 0.489 & 0.333 & 0.061 & 0.795 & 32.8 \\
\rowcolor{h2rfamilyone}  & LongCat & 0.412 & 0.502 & 0.208 & 0.000 & 0.799 & 27.9 & 0.394 & 0.476 & 0.306 & 0.032 & 0.798 & 31.2 \\
\rowcolor{h2rfamilyone} \multirow{-11}{*}{\rotatebox[origin=c]{90}{\textbf{F1 Rigid}}} & Wan2.2 & 0.377 & 0.547 & 0.229 & 0.000 & 0.776 & 28.5 & 0.422 & 0.553 & 0.249 & 0.000 & 0.774 & 29.8 \\
\rowcolor{h2rfamilytwo}  & Kling-V3 & 0.641 & 0.792 & 0.686 & 0.682 & 0.802 & 70.6 & 0.666 & 0.796 & 0.770 & 0.885 & 0.804 & 79.6 \\
\rowcolor{h2rfamilytwo}  & Seedance 2.0 & 0.760 & 0.847 & 0.737 & 0.742 & 0.797 & 76.4 & 0.735 & 0.854 & 0.828 & 0.927 & 0.801 & 84.5 \\
\rowcolor{h2rfamilytwo}  & Wan2.7 & 0.695 & 0.783 & 0.704 & 0.771 & 0.794 & 74.4 & 0.676 & 0.795 & 0.778 & 0.913 & 0.799 & 80.8 \\
\rowcolor{h2rfamilytwo}  & Grok Imagine & 0.646 & 0.729 & 0.491 & 0.527 & 0.791 & 59.1 & 0.683 & 0.770 & 0.573 & 0.344 & 0.807 & 57.4 \\
\rowcolor{h2rfamilytwo}  & Veo 3.1 & 0.670 & 0.754 & 0.508 & 0.049 & 0.785 & 45.9 & 0.691 & 0.824 & 0.620 & 0.151 & 0.793 & 53.8 \\
\rowcolor{h2rfamilytwo}  & Hunyuan-1.5 & 0.433 & 0.517 & 0.178 & 0.014 & 0.807 & 28.1 & 0.399 & 0.481 & 0.202 & 0.123 & 0.809 & 31.0 \\
\rowcolor{h2rfamilytwo}  & LTX-2.3 & 0.478 & 0.514 & 0.278 & 0.000 & 0.774 & 31.0 & 0.523 & 0.613 & 0.328 & 0.157 & 0.786 & 39.4 \\
\rowcolor{h2rfamilytwo}  & Mitty-EPIC & 0.515 & 0.588 & 0.537 & 0.579 & 0.746 & 57.5 & 0.507 & 0.609 & 0.540 & 0.393 & 0.746 & 52.2 \\
\rowcolor{h2rfamilytwo}  & SkyReels-V3 & 0.454 & 0.552 & 0.223 & 0.000 & 0.790 & 29.7 & 0.425 & 0.559 & 0.303 & 0.011 & 0.794 & 32.1 \\
\rowcolor{h2rfamilytwo}  & LongCat & 0.454 & 0.579 & 0.243 & 0.000 & 0.792 & 30.7 & 0.403 & 0.577 & 0.275 & 0.011 & 0.794 & 31.2 \\
\rowcolor{h2rfamilytwo} \multirow{-11}{*}{\rotatebox[origin=c]{90}{\textbf{F2 Mechanism}}} & Wan2.2 & 0.427 & 0.607 & 0.234 & 0.000 & 0.763 & 30.2 & 0.526 & 0.672 & 0.264 & 0.000 & 0.762 & 33.5 \\
\rowcolor{h2rfamilythree}  & Kling-V3 & 0.683 & 0.748 & 0.748 & 0.735 & 0.794 & 73.9 & 0.680 & 0.739 & 0.796 & 0.907 & 0.799 & 80.4 \\
\rowcolor{h2rfamilythree}  & Seedance 2.0 & 0.722 & 0.792 & 0.834 & 0.839 & 0.789 & 80.8 & 0.737 & 0.832 & 0.867 & 0.907 & 0.793 & 84.7 \\
\rowcolor{h2rfamilythree}  & Wan2.7 & 0.623 & 0.727 & 0.742 & 0.785 & 0.789 & 73.9 & 0.660 & 0.762 & 0.813 & 0.912 & 0.792 & 81.0 \\
\rowcolor{h2rfamilythree}  & Grok Imagine & 0.599 & 0.688 & 0.422 & 0.320 & 0.788 & 49.4 & 0.603 & 0.650 & 0.398 & 0.299 & 0.807 & 47.8 \\
\rowcolor{h2rfamilythree}  & Veo 3.1 & 0.687 & 0.764 & 0.504 & 0.072 & 0.780 & 46.9 & 0.619 & 0.770 & 0.597 & 0.240 & 0.791 & 53.8 \\
\rowcolor{h2rfamilythree}  & Hunyuan-1.5 & 0.462 & 0.448 & 0.214 & 0.000 & 0.803 & 28.1 & 0.409 & 0.469 & 0.193 & 0.022 & 0.805 & 27.7 \\
\rowcolor{h2rfamilythree}  & LTX-2.3 & 0.426 & 0.470 & 0.276 & 0.046 & 0.773 & 30.8 & 0.510 & 0.516 & 0.343 & 0.195 & 0.779 & 39.3 \\
\rowcolor{h2rfamilythree}  & Mitty-EPIC & 0.571 & 0.634 & 0.597 & 0.602 & 0.737 & 61.4 & 0.578 & 0.664 & 0.590 & 0.470 & 0.737 & 57.8 \\
\rowcolor{h2rfamilythree}  & SkyReels-V3 & 0.223 & 0.511 & 0.204 & 0.011 & 0.785 & 25.3 & 0.304 & 0.522 & 0.279 & 0.011 & 0.787 & 29.0 \\
\rowcolor{h2rfamilythree}  & LongCat & 0.482 & 0.586 & 0.238 & 0.000 & 0.786 & 31.0 & 0.365 & 0.474 & 0.271 & 0.000 & 0.789 & 28.6 \\
\rowcolor{h2rfamilythree} \multirow{-11}{*}{\rotatebox[origin=c]{90}{\textbf{F3 Insert}}} & Wan2.2 & 0.423 & 0.528 & 0.194 & 0.000 & 0.768 & 27.8 & 0.416 & 0.598 & 0.251 & 0.000 & 0.767 & 30.4 \\
\rowcolor{h2rfamilyfour}  & Kling-V3 & 0.795 & 0.862 & 0.847 & 0.690 & 0.773 & 78.7 & 0.804 & 0.857 & 0.892 & 0.853 & 0.776 & 85.0 \\
\rowcolor{h2rfamilyfour}  & Seedance 2.0 & 0.770 & 0.821 & 0.837 & 0.801 & 0.770 & 80.7 & 0.833 & 0.869 & 0.897 & 0.910 & 0.778 & 87.5 \\
\rowcolor{h2rfamilyfour}  & Wan2.7 & 0.799 & 0.818 & 0.852 & 0.778 & 0.775 & 80.9 & 0.789 & 0.824 & 0.877 & 0.893 & 0.777 & 85.1 \\
\rowcolor{h2rfamilyfour}  & Grok Imagine & 0.762 & 0.783 & 0.566 & 0.220 & 0.767 & 54.4 & 0.752 & 0.793 & 0.629 & 0.131 & 0.776 & 53.7 \\
\rowcolor{h2rfamilyfour}  & Veo 3.1 & 0.778 & 0.819 & 0.670 & 0.240 & 0.766 & 58.9 & 0.785 & 0.864 & 0.783 & 0.387 & 0.778 & 67.6 \\
\rowcolor{h2rfamilyfour}  & Hunyuan-1.5 & 0.608 & 0.634 & 0.138 & 0.000 & 0.796 & 30.7 & 0.566 & 0.644 & 0.164 & 0.013 & 0.800 & 31.5 \\
\rowcolor{h2rfamilyfour}  & LTX-2.3 & 0.485 & 0.595 & 0.334 & 0.000 & 0.753 & 33.8 & 0.540 & 0.615 & 0.462 & 0.102 & 0.760 & 41.8 \\
\rowcolor{h2rfamilyfour}  & Mitty-EPIC & 0.699 & 0.753 & 0.635 & 0.554 & 0.693 & 64.4 & 0.697 & 0.752 & 0.644 & 0.378 & 0.694 & 59.3 \\
\rowcolor{h2rfamilyfour}  & SkyReels-V3 & 0.540 & 0.716 & 0.292 & 0.000 & 0.774 & 35.3 & 0.491 & 0.695 & 0.448 & 0.048 & 0.774 & 40.4 \\
\rowcolor{h2rfamilyfour}  & LongCat & 0.409 & 0.596 & 0.250 & 0.000 & 0.778 & 30.4 & 0.408 & 0.514 & 0.349 & 0.045 & 0.783 & 33.5 \\
\rowcolor{h2rfamilyfour} \multirow{-11}{*}{\rotatebox[origin=c]{90}{\textbf{F4 Deform.}}} & Wan2.2 & 0.627 & 0.796 & 0.305 & 0.000 & 0.753 & 38.0 & 0.599 & 0.760 & 0.325 & 0.000 & 0.751 & 37.6 \\
\rowcolor{h2rfamilyfive}  & Kling-V3 & 0.659 & 0.796 & 0.742 & 0.749 & 0.804 & 74.6 & 0.667 & 0.820 & 0.803 & 0.901 & 0.809 & 81.5 \\
\rowcolor{h2rfamilyfive}  & Seedance 2.0 & 0.665 & 0.819 & 0.676 & 0.627 & 0.795 & 69.3 & 0.710 & 0.828 & 0.844 & 0.903 & 0.800 & 83.5 \\
\rowcolor{h2rfamilyfive}  & Wan2.7 & 0.680 & 0.808 & 0.769 & 0.766 & 0.795 & 76.3 & 0.690 & 0.826 & 0.841 & 0.913 & 0.796 & 83.3 \\
\rowcolor{h2rfamilyfive}  & Grok Imagine & 0.697 & 0.778 & 0.277 & 0.046 & 0.802 & 39.8 & 0.754 & 0.772 & 0.315 & 0.025 & 0.814 & 41.2 \\
\rowcolor{h2rfamilyfive}  & Veo 3.1 & 0.732 & 0.828 & 0.486 & 0.025 & 0.791 & 46.6 & 0.710 & 0.817 & 0.613 & 0.137 & 0.800 & 53.4 \\
\rowcolor{h2rfamilyfive}  & Hunyuan-1.5 & 0.620 & 0.704 & 0.206 & 0.018 & 0.811 & 34.7 & 0.557 & 0.666 & 0.188 & 0.011 & 0.815 & 32.5 \\
\rowcolor{h2rfamilyfive}  & LTX-2.3 & 0.449 & 0.541 & 0.236 & 0.000 & 0.786 & 29.8 & 0.522 & 0.656 & 0.353 & 0.091 & 0.788 & 38.9 \\
\rowcolor{h2rfamilyfive}  & Mitty-EPIC & 0.564 & 0.731 & 0.676 & 0.643 & 0.741 & 66.4 & 0.580 & 0.719 & 0.667 & 0.385 & 0.741 & 58.4 \\
\rowcolor{h2rfamilyfive}  & SkyReels-V3 & 0.427 & 0.627 & 0.278 & 0.016 & 0.793 & 32.6 & 0.421 & 0.634 & 0.344 & 0.024 & 0.795 & 34.8 \\
\rowcolor{h2rfamilyfive}  & LongCat & 0.413 & 0.564 & 0.240 & 0.000 & 0.796 & 29.8 & 0.437 & 0.572 & 0.287 & 0.010 & 0.800 & 32.0 \\
\rowcolor{h2rfamilyfive} \multirow{-11}{*}{\rotatebox[origin=c]{90}{\textbf{F5 Bulk}}} & Wan2.2 & 0.418 & 0.613 & 0.263 & 0.000 & 0.779 & 31.1 & 0.481 & 0.621 & 0.310 & 0.000 & 0.774 & 33.6 \\
\rowcolor{h2rfamilysix}  & Kling-V3 & 0.747 & 0.807 & 0.703 & 0.654 & 0.800 & 72.0 & 0.688 & 0.751 & 0.805 & 0.877 & 0.801 & 80.1 \\
\rowcolor{h2rfamilysix}  & Seedance 2.0 & 0.668 & 0.744 & 0.738 & 0.775 & 0.792 & 74.5 & 0.677 & 0.755 & 0.817 & 0.913 & 0.800 & 81.4 \\
\rowcolor{h2rfamilysix}  & Wan2.7 & 0.699 & 0.764 & 0.705 & 0.754 & 0.789 & 73.6 & 0.716 & 0.771 & 0.810 & 0.915 & 0.800 & 82.1 \\
\rowcolor{h2rfamilysix}  & Grok Imagine & 0.652 & 0.693 & 0.344 & 0.064 & 0.795 & 40.4 & 0.617 & 0.645 & 0.328 & 0.014 & 0.805 & 37.2 \\
\rowcolor{h2rfamilysix}  & Veo 3.1 & 0.731 & 0.825 & 0.456 & 0.010 & 0.784 & 45.2 & 0.706 & 0.793 & 0.582 & 0.138 & 0.794 & 52.0 \\
\rowcolor{h2rfamilysix}  & Hunyuan-1.5 & 0.517 & 0.503 & 0.182 & 0.000 & 0.807 & 28.8 & 0.502 & 0.572 & 0.157 & 0.052 & 0.808 & 30.5 \\
\rowcolor{h2rfamilysix}  & LTX-2.3 & 0.574 & 0.600 & 0.284 & 0.007 & 0.766 & 34.0 & 0.572 & 0.592 & 0.355 & 0.076 & 0.776 & 38.2 \\
\rowcolor{h2rfamilysix}  & Mitty-EPIC & 0.615 & 0.709 & 0.601 & 0.557 & 0.740 & 62.0 & 0.632 & 0.721 & 0.558 & 0.318 & 0.740 & 54.0 \\
\rowcolor{h2rfamilysix}  & SkyReels-V3 & 0.636 & 0.725 & 0.272 & 0.000 & 0.785 & 36.4 & 0.631 & 0.740 & 0.338 & 0.000 & 0.787 & 38.6 \\
\rowcolor{h2rfamilysix}  & LongCat & 0.662 & 0.671 & 0.278 & 0.000 & 0.790 & 36.2 & 0.559 & 0.654 & 0.299 & 0.022 & 0.791 & 35.7 \\
\rowcolor{h2rfamilysix} \multirow{-11}{*}{\rotatebox[origin=c]{90}{\textbf{F6 Surface}}} & Wan2.2 & 0.679 & 0.746 & 0.319 & 0.000 & 0.772 & 38.7 & 0.630 & 0.716 & 0.314 & 0.000 & 0.771 & 37.3 \\
\bottomrule
\end{tabular}
\caption{Main-evaluation metric breakdown by task family for all 11 evaluated models. Parallel-Jaw Gripper and Dexterous Hand targets are reported separately. Video Quality is the task-family mean of M5 and contributes 0.10 to H2RCore. Overall values in Table~\ref{tab:main_results_by_embodiment} are computed from the unrounded per-video scores.}
\label{tab:family_breakdown}
\end{table*}

\begin{table*}[!t]
\centering
\footnotesize
\renewcommand{\arraystretch}{1.04}
\setlength{\tabcolsep}{4pt}
\begin{tabular*}{\textwidth}{@{\extracolsep{\fill}}p{0.14\textwidth}p{0.25\textwidth}p{0.56\textwidth}@{}}
\toprule
Metric & Evidence & Computation \\
\midrule
M1: Goal & 25 generated frames and final-state predicates. & Weighted mean of normalized 0--4 predicate scores. \\
M2: Action & 25 generated frames and required events. & Weighted mean of normalized event-completion scores. \\
M3: Contact & 25 source frames, 25 generated frames, and the contact specification. & Mean of applicable contact dimensions; zero when source grounding fails. \\
M4: Embodiment & 25 generated frames and the target specification. & Weighted mean of five embodiment dimensions; zero on a hard failure. \\
M5: Quality & Decoded generated video. & Mean of MUSIQ, CLIP aesthetic, temporal-stability, and AMT-S scores. \\
\bottomrule
\end{tabular*}
\caption{Metric implementation summary. Detailed definitions, evidence rules, diagnostic components, and per-video formulas are given in Section~\ref{sec:detailed_evaluation}.}
\label{tab:h2r_metrics}
\end{table*}

\section{Limitations}
\label{sec:limitations}

H2R-Bench evaluates visible evidence of human-to-robot transfer, not physical executability or downstream policy performance. Its 120 EgoDex sources and two target embodiments cover only part of the variation found in manipulation settings. The current benchmark is also limited to short clips; extending it to longer demonstrations may benefit from efficient spatiotemporal modeling~\cite{yang2025moma}. The native-interface comparison also combines model capability with differences in source-conditioning interfaces. Finally, sampled visual evidence and MLLM judgments can remain uncertain under occlusion, subtle contact, or severe generation artifacts despite multi-judge aggregation and human validation. Future extensions could use dense visual grounding for ambiguous task entities~\cite{ma2022fusioner,ma2023attrseg,mao2025safire,yang2024multimodal,yang2024remamber,yang2026genmask}. Geometry-aware novel-view synthesis may provide complementary visual evidence~\cite{shi2022darf}, and fine-grained 3D reconstruction may support object-geometry diagnostics~\cite{shi2026cei}.

\section{Human Evaluation Interface}
\label{sec:human_evaluation_interface}

\begin{figure*}[!t]
\centering
\includegraphics[width=0.94\textwidth]{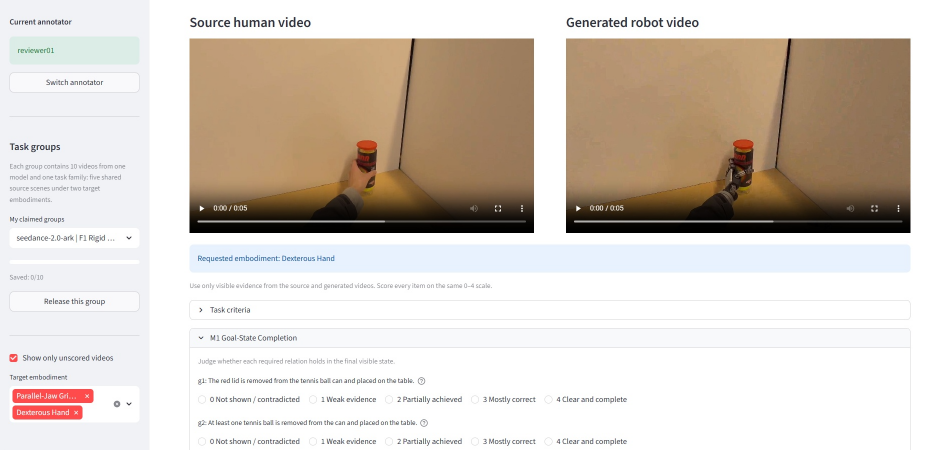}
\caption{Human-evaluation interface for H2R-Bench. The page pairs each source demonstration with its generated robot video and presents the requested embodiment and expandable M1--M4 criteria. The sidebar records task assignment and progress over balanced ten-video groups.}
\label{fig:human_evaluation_interface}
\end{figure*}

Figure~\ref{fig:human_evaluation_interface} shows the interface used to collect human M1--M4 scores. Each task group contains five shared source scenes from one model and task family under both target embodiments, yielding ten videos. Annotators inspect the source and generated clips side by side, verify the requested embodiment, and score the source-derived criteria on a common 0--4 scale. Scores are stored separately for each annotator, and automatic MLLM judgments are not displayed.

\begin{figure*}[!t]
\centering
\begingroup
\setlength{\fboxsep}{9pt}%
\fcolorbox{h2rpromptouterborder}{h2rpromptouterbg}{%
\begin{minipage}{0.94\textwidth}
\centering
{\large\bfseries Generation Prompts and Structured Annotation Input}\par\medskip

\begin{minipage}[t]{0.485\linewidth}
\vspace{0pt}
\promptcard{h2rpromptvideoaccent}{h2rpromptvideobg}{Video-Conditioned Generation}{%
\textbf{Input:} complete source demonstration video.\newline
[VIDEO EDIT]\newline
Generate a 5-second edit from the reference clip. Keep the source camera, scene, lighting, background, object identities, layout, required actions, contact timing, and visible final state. Replace visible human hands, wrists, sleeves, and forearms with continuous robotic arms ending in five-finger dexterous hands. Objects may move only through visible robotic contact, stable support, or physically plausible pushing. Avoid scene or object substitution, residual human hands, invented actions, floating objects, missing contact, unstable final states, and warped robot geometry.}
\end{minipage}\hfill
\begin{minipage}[t]{0.485\linewidth}
\vspace{0pt}
\promptcard{h2rpromptimageaccent}{h2rpromptimagebg}{Frame-Conditioned Generation}{%
\textbf{Input:} first and last source frames in temporal order.\newline
[IMAGE-CONDITIONED VIDEO GENERATION]\newline
Use the input frames as visual boundary conditions for the same manipulation. Preserve the source camera, scene, lighting, background, object identities, layout, required actions, and visible final state. Replace visible human hands, wrists, sleeves, and forearms with continuous robotic arms ending in five-finger dexterous hands. Maintain visible robot--object contact and physically plausible object motion. Avoid residual human hands, scene changes, invented actions, floating objects, missing contact, unstable final states, and warped robot geometry.}
\end{minipage}

\medskip

\begin{minipage}[t]{0.485\linewidth}
\vspace{0pt}
\promptcard{h2rpromptembaccent}{h2rpromptembbg}{Embodiment-Specific Replacement}{%
\textbf{Parallel-Jaw Gripper:} Replace visible human manipulators with continuous robotic arms ending in rigid two-jaw parallel grippers. The jaws open and close laterally. Do not show human skin, gloves, soft or dexterous fingers, extra fingers, or claw grippers.\newline\newline
\textbf{Dexterous Hand:} Replace visible human manipulators with continuous robotic arms ending in five-finger dexterous hands, each with one opposable thumb and four articulated mechanical fingers. Do not show human skin, gloves, parallel-jaw or claw grippers, missing fingers, or extra fingers.}
\end{minipage}\hfill
\begin{minipage}[t]{0.485\linewidth}
\vspace{0pt}
\promptcard{h2rpromptannaccent}{h2rpromptannbg}{Structured Source-Video Annotation}{%
\textbf{Input:} source video, 32 uniformly sampled frames, and the JSON schema.\newline
[STRUCTURED SOURCE-VIDEO ANNOTATION]\newline
Use only visible evidence and record uncertainty instead of inferring occluded contact or unseen state changes. Identify the task, scene, initial and final states, manipulated entities, and task family. For each required event, record its temporal span, hand roles, contact mode and region, state transition, success evidence, and uncertainty. Select keyframes for the initial state, first contact, core transition, and final state. Return valid JSON containing transfer requirements, permissible adaptations, and invalid outcomes for both target embodiments.}
\end{minipage}

\medskip

\begin{minipage}[t]{0.485\linewidth}
\vspace{0pt}
\promptcard{h2rpromptjudgeaccent}{h2rpromptjudgebg}{M3 Judge Prompt}{%
\textbf{Role:} strict visual evaluator for Functional Contact Transfer.\newline
\textbf{Evidence:} source frames 1--25 and generated frames 1--25, numbered independently.\newline
Judge only visible evidence. Do not infer occluded contact or unseen state changes, and use the case specification as a checklist rather than evidence. First apply the source-grounding gate: fail only for substantial scene or task-entity substitution. If grounding passes, score contact-region transfer, contact establishment, manipulation-mode transfer, temporally supported object response, and embodiment-compatible contact strategy on the 0--4 rubric. The robot need not copy the human pose or trajectory. Do not score final-goal completion, action coverage, generic robot appearance, or video quality. Return one valid JSON object with concise evidence-frame references and reasons.}
\end{minipage}\hfill
\begin{minipage}[t]{0.485\linewidth}
\vspace{0pt}
\promptcard{h2rpromptoutputaccent}{h2rpromptoutputbg}{Structured Judge Output}{%
\{\newline
~~"source\_grounding": \{\newline
~~~~"pass": true,\newline
~~~~"evidence\_source\_frames": [\ldots],\newline
~~~~"evidence\_generated\_frames": [\ldots],\newline
~~~~"reason": "\ldots"\newline
~~\},\newline
~~"dimensions": \{\newline
~~~~"contact\_region\_transfer": \{\ldots\},\newline
~~~~"contact\_establishment": \{\ldots\},\newline
~~~~"manipulation\_mode\_transfer": \{\ldots\},\newline
~~~~"temporally\_supported\_object\_response": \{\ldots\},\newline
~~~~"embodiment\_compatible\_contact\_strategy": \{\ldots\}\newline
~~\}\newline
\}\newline
Each applicable dimension contains \{"applicable": true, "score\_0\_to\_4": 0, source/generated evidence frames, and a concise reason\}.}
\end{minipage}
\end{minipage}%
}
\endgroup
\caption{Prompt formats used for generation, source-video annotation, and MLLM judging. The first row contrasts video- and frame-conditioned generation instructions; the second shows the embodiment clause and evaluation-only annotation input; and the third gives an abridged M3 judge prompt with its structured output.}
\label{fig:prompt_examples}
\end{figure*}

\raggedbottom

\end{document}